\documentclass[twoside]{article}

\usepackage[accepted]{aistats2024}

\usepackage[utf8]{inputenc} 
\usepackage[T1]{fontenc}    
\usepackage{lipsum}
\usepackage{tabularx}
\usepackage{enumitem}
\usepackage{adjustbox}
\usepackage[hidelinks,colorlinks=true,linkcolor=blue,citecolor=blue,hyperfootnotes=false]{hyperref}
\usepackage{comment}
\usepackage{Definitions}
\usepackage{pdfpages}
\usepackage[ruled,vlined,boxed,linesnumbered]{algorithm2e}
\usepackage{url}            
\usepackage{booktabs}       
\usepackage{amsfonts}       
\usepackage{nicefrac}       
\usepackage{microtype}      
\usepackage{xcolor}         
\usepackage{mathtools}
\usepackage{color}

\usepackage[ruled,vlined]{algorithm2e}
\usepackage{algorithmic}
\usepackage[hidelinks,colorlinks=true,linkcolor=blue,citecolor=blue]{hyperref}

\usepackage{Definitions}
\usepackage{wrapfig}
\usepackage{graphicx}  
\usepackage{float}
\usepackage{bbm}    
\usepackage{amsmath}
\usepackage{empheq}
\usepackage{diagbox,multirow,multicol}
\newcommand{\tabincell}[2]{\begin{tabular}{@{}#1@{}}#2\end{tabular}}
\usepackage{makecell}
\usepackage{natbib}
\setcitestyle{numbers,square}

\usepackage{fancyhdr}
\usepackage{titlesec}
\titlespacing*{\section}
{0pt}{1ex plus 1ex minus .2ex}{0.5ex plus .2ex}
\titlespacing*{\subsection}
{0pt}{1ex plus 1ex minus .2ex}{0.5ex plus .2ex}
\titlespacing*{\subsubsection}
{0pt}{1ex plus 1ex minus .2ex}{0.5ex plus .2ex}
\setlist[enumerate]{itemsep=0ex,topsep=0ex}
\setlist[itemize]{itemsep=0ex,topsep=0ex}

\begin{document}

%

%

\twocolumn[

\aistatstitle{Unveiling Latent Causal Rules: A Temporal Point Process Approach for Abnormal Event Explanation}

\aistatsauthor{ $\text{Yiling Kuang}^{1, *}$ \And $\text{Chao Yang}^{1, *}$ \And  $\text{Yang Yang}^1$ \And $\text{Shuang Li}^{1,\dagger}$}

\aistatsaddress{ $^1$The Chinese University of Hong Kong, Shenzhen } ]

\begin{abstract}
  In high-stakes systems such as healthcare, it is critical to understand the causal reasons behind unusual events, such as sudden changes in patient's health. Unveiling the causal reasons helps with quick diagnoses and precise treatment planning. In this paper, we propose an automated method for uncovering ``if-then'' logic rules to explain observational events. We introduce {\it temporal point processes} to model the events of interest, and discover the set of latent rules to explain the occurrence of events. To achieve this goal, we employ an Expectation-Maximization (EM) algorithm. In the E-step, we calculate the posterior probability of each event being explained by each discovered rule. In the M-step, we update both the rule set and model parameters to enhance the likelihood function's lower bound. Notably, we will optimize the rule set in a {\it differential} manner. Our approach demonstrates accurate performance in both discovering rules and identifying root causes. We showcase its promising results using synthetic and real healthcare datasets.
\end{abstract}

\section{Introduction}

Detecting and understanding abnormal events is crucial in various fields. In healthcare, quickly spotting and diagnosing sudden patient deterioration can save lives. In e-commerce, detecting odd user behavior may stop fraud and ensure safe shopping experiences. In manufacturing, finding irregularities in production processes can make operations more efficient and reduce downtime. These examples highlight the importance of abnormal event detection for enhancing safety, security, and efficiency in various areas.

However, effectively implementing abnormal event detection comes with challenges. A  major obstacle is the lack of labeled abnormal event data during training. Labeling events as normal or abnormal is often expensive and labor-intensive.  Consider the context of credit card transactions. When the system encounters an unusually large transaction, determining its underlying rationale can be intricate. It may indeed be a legitimate and routine transaction, or it could be indicative of fraudulent activity. Identifying the true reasons requires thorough investigation, which is both time- and resource-consuming. In our paper, we address this issue by introducing a {\it rule-based probabilistic} approach for predicting and explaining events. We treat the event's explanatory reasons as hidden variables and use an EM algorithm to learn rules that explain events without explicit label information regarding anomaly. For each observed event, the triggered rules will provide insight into the most probable causal factors and determine whether the event should be flagged as abnormal or normal. 

We introduce {\it temporal point process} to model events of interest in continuous time. The {\it time-to-event} is treated as a random variable characterized by the {\it intensity} function, indicating the occurrence rate of the event in real-time. We formulate the intensity function as a  {\it mixture model}, where each component is defined by an ``if-then'' rule factor. We assume that the ``if'' condition should occur before the ``then'' part, and the ``if'' condition should also specify the temporal order constraints that the relevant logic variables must meet, such as ``if $X_1$ and $X_2$ are true and $X_1$ happens before $X_2$, then $Y$ is true.''  These explanatory temporal logic rules will be discovered from data, which {\it compete} to influence when an event occurs. When an event happens, the specific rule that is triggered based on preceding events will provide an explanation for its occurrence. 

As mentioned above, we formulate an EM algorithm to simultaneously learn the rule set and temporal point processes model parameters. In the E-step, we calculate the assignment probability of each event to its causal rules based on our current estimations of the model parameters and the rule set. In the M-step, we update the rule set and model parameters to maximize the expected likelihood calculated in E-step. 

Notably, {\it learning the rule set} poses a formidable challenge due to its inherently {\it combinatorial} nature. It requires selecting the proper subsets of logic variables and then considering their temporal order relations to create a logic rule. This variable selection process is both time-consuming and non-differentiable. In our paper, we propose to solve a relaxed continuous optimization problem by directly modeling the logic variable selection probability in forming a rule. Specifically, each rule is generated by {\it a reparameterized K-subset sampling with the Gumbel noise injected to make the rule sampling process differentiable}~\citep{xie2019reparameterizable}. We further use a dummy predicate trick to accommodate for various rule lengths, while still being limited by $K$. In the M-step, both the rule set and the model parameters can be learned end-to-end in a differentiable way. 

Our contributions can be summarized as follows.

\begin{itemize}
    \item We leverage the continuous-time temporal point process modeling framework and formulate the intensity function as a rule-informed mixture model. Our rule-based probabilistic model is inherently interpretable, ensuring safe and reliable event prediction and explanation. 
    \item We design an EM learning algorithm that can accommodate the latent logical reasons for each event. Our overall learning algorithm is efficient and differentiable, which can automatically discover the set of explanatory logic rules from the population data and determine the most likely logical reason for each event. 
    \item We conduct empirical evaluations of our algorithm using both synthetic and real datasets, demonstrating its strong performance in rule discovery and root cause determination.
\end{itemize}

\setlength{\abovedisplayskip}{1pt}
\setlength{\abovedisplayshortskip}{0pt}
\setlength{\belowdisplayskip}{2pt}
\setlength{\floatsep}{2ex}
\setlength{\textfloatsep}{2ex}
\setlength{\parskip}{0.5ex}

\section{Related Work}
We will compare our method with some existing works from the following aspects.

\textbf{Temporal Point Process (TPP) Models} 
TPP models provide an elegant tool for modeling the interevent time intervals or time-to-event in continuous time. Current research in this area primarily focuses on enhancing the flexibility of intensity functions. For example, \citet{du2016recurrent} introduced the RMTPP, a neural point process model that employs a Recurrent Neural Network (RNN) to model the intensity function. \citet{mei2017neural} further improved RMTPP by developing a continuous-time RNN. \citet{zuo2020transformer} and \citet{zhang2020self} leveraged the self-attention mechanism to capture the long-term dependencies of events. However, understanding how these flexible black-box models work, which excel at event prediction with sufficient training data, presents challenges. It is difficult to interpret the underlying mechanisms driving the occurrence of events from these models. In scenarios like abnormal event detection, the objective goes beyond merely predicting unusual events; it also involves providing causal explanations for these occurrences. The absence of interpretability renders black-box models unsuitable for high-stakes systems where transparency is critical.

Recently, some researchers in AI started emphasizing the importance of developing inherently interpretable models rather than solely relying on explaining black-box models~\citep{rudin2019stop}. Along this line of research, \cite{li2020temporal,li2021explaining} designed a rule-informed intensity function for TPPs. The highly structured intensity function, encoding logical rules as prior knowledge, enables the discovery of logic rules to explain events. Furthermore, these models can perform effectively in small data regimes. However, their approach tends to discover global explanatory logic rules from population data to maximize likelihood~\citep{li2021explaining}, while paying less attention to identifying the most probable explanatory logic rules at the individual event level, unlike our method. Additionally, their rule learning algorithm is non-differentiable, relying on a branch-and-bound approach to progressively add rules to the rule set.

\textbf{Rule Set Mining} The Discovery of explanatory rule sets from event sequences in an unsupervised manner has long been an interesting topic. Traditional approaches, such as Itemset Mining Methods like Apriori~\citep{agrawal1994fast} and NEclatclosed~\citep{aryabarzan2021neclatclosed}, focus on identifying frequent event items from population data to generate association rules but overlook event order. Sequential Patterns Mining methods like CM-SPADE~\citep{fournier2014fast} and VGEN~\citep{fournier2014vgen}, aim to uncover temporal relationships from event sequences but often lack precision and struggle with the fine-grained recorded timestamp information. In contrast to these unsupervised methods, supervised learning methods such as Inductive Logic Programming (ILP)~\citep{srinivasan2001aleph} offer a principled rule learning approach, albeit usually requiring balanced positive and negative examples for optimal performance. The rule set is discovered under the principle that all positive samples are accounted for by at least one rule, while none of the negative samples are covered by the rule set.  As a comparison to all previous methods, our rule discovery method based on TPPs can naturally deal with fine-grained temporal information of events, and the learning algorithm is unsupervised, without the need for labeled positive and negative event samples.

\section{Problem Setup}
We define a set of logic variables (or predicates) as $\Xcal$, where each variable $X_u \in \Xcal$ is a boolean variable. Our target event, such as a sudden change in a patient's health or an unusually large transaction, is denoted as $Y$ and we also let $Y \in \Xcal$. 

Adding a temporal dimension to the logic variables, we assume that each grounded logic variable from the observations is a sequence of spiked events, denoted as $\{X_u(t)\}_{t\geq 0}$, where each $X_u(t) \in \{0, 1\}$, $\forall t$. Specifically, $X_u(t)$ jumps to 1 (i.e., true) at the time step when the event happens. In our problem, we consider the grounded logic variables as degenerated events, and each data sample observed up to time $t$ is represented as a $|\Xcal|$-dimensional multivariate TPP, denoted as $\mathcal{H}_t= \{X_u(t)\}_{u=1, \dots, |\Xcal|}$. 

We are interested in modeling and learning logic rules from history $\mathcal{H}_t$ to provide insight into the occurrence of $\{Y(t)\}_{t\geq 0}$. We consider a continuous-time setting, where the distribution of the duration until event $Y$ happens is characterized by the {\it conditional intensity function}, denoted as $\lambda(t\mid\Hcal_{t-})$. Here, the intensity function is conditional on history, and $t-$ means up to $t$ but not including $t$. By definition, 
\[
  \lambda(t\mid\Hcal_{t-})dt=\mathbb{E}[N([t, t+dt))\mid \Hcal_{t-}],
\]
where $N([t, t+dt))$ denotes the number of events occurring in the interval $[t, t+dt)$. Given the occurrence time of event $Y$, such as $(t_1, \dots, t_n)$, the joint likelihood function of these events is given by 
\begin{align}
&\quad p(t_1, \dots, t_n)= \prod_{i=1}^n p^*(t_i), \quad \text{with} \nonumber\\
& p^*(t_i) = \lambda^* \left(t_i \right)\exp \left(-\int_{t_{i-1}}^{t_i} \lambda^*(s )ds \right),
\label{eq:likelihood}
\end{align}

where we denote $p^*(t):= p(t \mid \Hcal_{{t}-})$ and $ \lambda^* (t):= \lambda (t \mid \Hcal_{t-})$ to simplify the notation. 

In this paper, we will model $\lambda^*(t)$ using the rule-informed features, and through learning the rules we aim to understand the underlying logical reasons to explain the occurrence of target events. Specifically, our goals include:

({\it i}) Discover the rule set $\mathcal{F}:=\{f_1, f_2, ..., f_H\}$ from the entire events $\mathcal{H}_{t_n}$ up to $t_n$, with each rule having a general Horn clause form 
    {\small \begin{align}
    f: \quad   Y \gets \left(\bigwedge \limits_{X_u \in \mathcal X_f} X_u \right)\bigwedge \left(\bigwedge \limits_{X_u, X_v \in \mathcal X_f} R_j(X_u,X_v) \right),
    \label{eq:rule}
    \end{align}}
    
where $\mathcal X_f$ is the set of the involved body predicates in the rule $f$. ${R_j(X_u, X_v)}$ is the temporal relation between the selected predicates $X_u$ and $X_v$. $R_j$ can take any of the four types, i.e., ``Before'', ``Equal'', ``After'', and ``None'', specifying the temporal order constraints for the predicate pair. Notably, the temporal relations can be none, meaning there are no temporal order constraints. 

({\it ii}) Infer the assignment of each occurred target event $\{Y(t_1), \dots, Y(t_n)\}$ to the following two groups:
\begin{itemize}
    \item spontaneous events (i.e., cannot be explained by any logic rule or it is meaningless to be explained);
    \item can be explained by one logic rule from the discovered rule set $\mathcal{F} = \{f_1, f_2, ..., f_H\}$ and our algorithm will determine the most probably one.
\end{itemize}

\section{Model}
\begin{figure}[ht]
\centering
\includegraphics[width=0.45\textwidth]{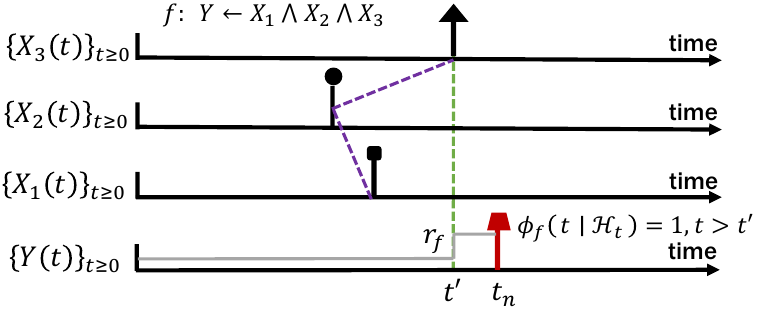}
\vspace{-15pt}
\caption{Illustration of the feature construction. For a logic rule $f$ such as: {\small $Y \gets X_1\bigwedge X_2\bigwedge X_3$}, whenever the body condition becomes true, the rule gets fired. As a result, the feature $\phi_f$ becomes 1, and the intensity function of the head predicate is boosted. }
\label{feature}
\end{figure}

We first describe how to construct the intensity function for $\{Y(t)\}_{t\geq 0}$ based on the logic-informed features. For a general rule as defined in Eq.(\ref{eq:rule}), we can construct the rule-informed feature as 
 \vspace{5pt}
{\small \begin{align} 
\phi_{f}(\Hcal_t) = \prod_{X_u \in \mathcal X_f } X_u(t_u)\cdot\prod_{X_u, X_v \in \mathcal X_f} R_j(t_u,t_v) \in \{0,1\}
\label{eq:feature}
\end{align}}

where we need to check from $\Hcal_t$ whether each $X_u \in \mathcal X_f $ has been once grounded as true at $t_u$ where $t_u < t$, and whether the temporal relation holds, which is grounded by their times (for multiple events we choose the last one). The temporal relation $R_j(t_u,t_v)$ is grounded according to its type, ``Before'', ``Equal'', ``After'', or ``None'' (denoted as $R_b, R_e, R_a, R_n$ respectively) by the event times as follows,
{$$ 
\begin{aligned}
R_b\left(t_u, t_v\right) & =\mathbb{I}\left\{t_u-t_v< -\delta\right\} \\
R_e\left(t_u, t_v\right) & =\mathbb{I}\left\{\left|t_u-t_v\right|\leq \delta \right\} \\
R_a\left(t_u, t_v\right) & =\mathbb{I}\left\{t_u-t_v>\delta \right\}\\
R_n\left(t_u, t_v\right) & = 1
\end{aligned}
$$}

where $\delta$ (can be 0) is specified as the time tolerance. If the body condition of $f$ is grounded to be true given history, then the feature $\phi_{f}(\Hcal_t) $ is 1; otherwise it is 0. A better illustration of how to construct the feature can be found in  Fig.\ref{feature}. 

Given the rule-informed feature defined in Eq.(\ref{eq:feature}), we can build our intensity model for $\{Y(t)\}_{t\geq 0}$ as follows. For the $i$-th event in the dataset, by first assuming all the rules $\mathcal{F} = \{f_1, f_2, ..., f_H\}$ are known, we model the intensity function as a {\it mixture-of-components} form,
\vspace{5pt}
{\small \begin{align}
\lambda^*(t_i \mid  \bm{z}_i) = z_{i0} b_0 + \sum_{h=1, \dots, H} \gamma_h z_{ih} \phi_{f_h} (\Hcal_{t_i})
\label{eq:intensity}
\end{align}}

where $b_0$ is the base term, $h$ is the index of the rule, and $\gamma_{h}>0$ denotes the  impact weight of each rule $f_h$. The {\it latent} variable $\bm{z}_{i}=[z_{ih}]_{h = 0, 1, \dots, H}$ is a {\it one-hot} vector. When $z_{ih}=1$, $h\neq 0$, it means event $i$ is due to the $h$-th rule factor, which provides logical reasons to explain its occurrence. We use $z_{i0}$ to indicate that {\it the event cannot be explained by any discovered rule}. We assume that each categorical $\bm{z}_i$ has the same prior distribution $\bm{\pi}=[\pi_h]_{h = 0, 1, \dots, H}$, which indicates the probability of each component appearing in the population and is our learnable parameter.

In the next section, we design an EM algorithm, which aims to jointly learn the rule set and the TPP model parameters from data. Meanwhile, in the E-step, we perform the inference of each $\bm{z}_i$, which indicates the most probably explanatory rule for event $i$.

\section{Learning: EM Algorithm}
To accommodate the {\it latent} rule assignment $\bm{z}_i$ for each event, it is natural to use an EM learning algorithm.
\begin{itemize}
\item [({\it i})]In the E-step, the {\it posterior probability} over $\bm{z}_i=[z_{ih}]_{h=0, \dots, H}$ for each event $Y(t_i)$ is computed.   
\item [({\it ii})]In the M-step, both the rule set and the point process model parameters are updated by maximizing the {\it expected likelihood} found in E-step. Denote all the learnable model parameters as
{ $$\theta = [b_0,[\gamma_h]_{h \in [1,\dots,H]}, \bm{\pi}, \mathcal{F}].$$}
\end{itemize}
Notably, we devise a series of techniques in EM to enhance the stability of the optimization process. The E-step can be successfully resolved in a closed-form solution. The M-step is more involved, for which we employ three-fold approximations and reparameterization to facilitate the end-to-end differentiable training, which will be discussed later. 

\textbf{Complete Data Likelihood} To simplify the derivation, let's first focus on the $i$-th event $Y(t_i)$ and consider its complete data likelihood. The following derivations can be easily extended to the entire event sequence.  

For the $i$-th event $Y(t_i)$, given the rule index $\bm{z}_i=[z_{ih}]_{h = 0, \dots, H}$, we can write the {\it likelihood} for the {\it complete data} $\{(t_i, \bm{z}_i)\}$ as
{\small \begin{align}
& p^*_{\theta}(t_i , \bm{z}_i) = p^*_{\theta}(t_i \mid \bm{z}_i) p( \bm{z}_i ), \quad \text{where} \nonumber \\
& p(z_{ih}=1)= \pi_{h}, \quad \text{with}\quad 0\leq\pi_{h}\leq 1, \, \sum_{h=0}^H\pi_h=1 \nonumber \\
& p^*_{\theta}(t_i \mid z_{ih}=1)\nonumber \\
&=\lambda^*_{\theta}(t_i \mid z_{ih}=1) \cdot \exp \left( -\int_{t_{i-1}}^{t_i} \lambda^*_{\theta}(s \mid z_{ih}=1)ds\right).
\label{eq:complete_like}
\end{align}}

The complete data log-likelihood $\log p^*_{\theta}(t_i , \bm{z}_i)$ is easier to optimize compared to the log-likelihood of data, computed as
{\small \begin{align}
\ell (\theta) = \log \left[ \sum_{  \bm{z}_i} p_{\theta}^*( t_i , \bm{z}_i)\right]
\end{align}}

which requires the marginalization of the latent variable $\bm{z}_i$.
\subsection{E-step: Compute Posterior  \texorpdfstring{$Q^{new}(\bm{z}_i)$}{Q}}
To derive the lower bound of the log-likelihood, let's denote the distribution for $\bm{z}_i=[z_{ih}]_{h=0,1, \dots, H}$ as $Q(\bm{z}_i)$ and $Q\left(\bm{z}_i\right) \geq 0$ and $\sum_{\bm{z}_i}Q\left(\bm{z}_i\right) =1$. The above constraints make $Q(\bm{z}_i)$ a valid probability mass function. 

Now we can create a lower bound for the log-likelihood function 
{\small \begin{align*}
\ell (\theta)= \log \mathbb{E}_{ Q(\bm{z}_i)} \left [ \frac{p^*_{\theta}(t_i, \bm{z}_i )}{Q(\bm{z}_i)} \right ] 
\geq  \mathbb{E}_{ Q(\bm{z}_i)}\left[\log \frac{p_{\theta}^*\left(t_i, \bm{z}_i\right)}{Q\left(\bm{z}_i\right)}\right]
\end{align*} }

where the inequality is due to the Jensen's inequality. 

In the current iteration of the EM algorithm, suppose we aim to find $Q^{new}(\bm{z}_i)$, so that the lower bound equals to log-likelihood $\ell\left(\theta^{old} \right)$. According to Jensen's inequality, the equality between the lower bound and $\ell\left(\theta^{old} \right)$ holds if $\frac{p_{\theta^{old}}^*\left(t_i, \bm{z}_i \right)}{Q^{new}\left(\bm{z}_i\right)}$ is a constant. In other words, $Q^{new}(\bm{z}_i)$ needs to take the posterior of $\bm{z}_i$ to make the lower bound tight, i.e., 
{\begin{align}
\textbf{E-step:}\quad & Q^{new}\left(z_{ih}=1\right)  =\frac{p^*_{\theta^{old} }\left(t_i, z_{ih}=1\right)}{\sum_{h'=0}^{H} p^*_{\theta^{old}}\left(t_i, z_{ih'}=1\right)} \nonumber \\
& =\frac{\pi_h^{old} p^*_{\theta^{old}}(t_i \mid z_{ih}=1) }{\sum_{h'=0}^{H}  \pi^{old} _{h'}p^*_{\theta^{old}}(t_i \mid z_{ih'}=1)}
\label{eq:Q_iter}
\end{align}}

with a more detailed calculation can be found in Eq.(\ref{eq:complete_like}).
\subsection{M-step: Optimize \texorpdfstring{$\theta^{\text{new}}$}{theta new}}
For the M-step, we aim to update the model parameters $\theta$ (including the rule set $\Fcal$) to maximize the expected log-likelihood computed in the E-step, which is a lower bound of the log-likelihood, i.e., 
{\begin{align*}
\theta^{new}:=\arg \max _{\theta} \underbrace{  \mathbb{E}_{ Q^{new}( \bm{z}_i )}\left[\log \frac{p_{\theta}^*\left(t_i, \bm{z}_i  \right)}{Q^{new}\left( \bm{z}_i \right)}\right]}_{\text{``energy''}}
\end{align*}}

which is equal to solve (since $Q^{new}( \bm{z}_i ) $ does not depend on $\theta$)
{\begin{align}
& \theta^{new}:=\arg \max _{\theta}  \mathbb{E}_{Q^{new}( \bm{z}_i )}  \left[\log p_{\theta}^*\left(t_i, \bm{z}_i \right)\right]
\end{align}}

where $\log p_{\theta}^*\left(t_i, \bm{z}_i \right)$ is the complete data log-likelihood. We further partition $\theta$ as $\theta :=[\theta_0, \Fcal]$ where $\theta_0$ contains the continuous model parameters excluding $\Fcal$. Then we rewrite the above formulation as 
\begin{align}
 \max_{\theta_0} \max_{\Fcal}  \mathbb{E}_{Q^{new}( \bm{z}_i )}  \left[\log p^*_{\theta_0, \Fcal}\left(t_i, \bm{z}_i\right)\right]
\end{align}
and we will alternate the optimization of $\theta_0$ and $\Fcal$. 
\subsection{Optimization w.r.t. \texorpdfstring{$\mathcal{F}$}{F}}
We first discuss how to update $\Fcal$ given current $\theta_0.$ We aim to propose a differentiable method to learn the rule set. To achieve this goal, we will employ three-fold approximations and reparameterization: 
\begin{itemize}
\item [({\it i})] {\bf Continuous Relaxation for Rule Learning:} We first formulate the rule learning problem as a logic variable selection problem and encode the selection results into a binary matrix $A\in \{0, 1\}^{H \times (|\Xcal|+M)}$. We then perform the continuous relaxations and relax $A$ to the selection probability $\tilde{A}\in [0, 1]^{H \times (|\Xcal|+M)}$. 
\item [({\it ii})] {\bf Continuous Approximation of the Boolean Logic Features:} We employ a logic-informed boolean feature approximation to ensure that the log-likelihood becomes a deterministic function of parameter $\tilde{A}$.
\item [({\it iii})] {\bf Differentiable Top-$K$ Subset Sampling:} 
We conduct a differentiable top-$K$ subset sampling by injecting Gumbel noise into $\tilde{A}$ to generate each rule. This approach introduces noise during rule generation, promoting exploration and aiding in avoiding suboptimal solutions. To control the generated rule length, we further choose to use a Top-$K$ subset sampling~\citep{xie2019reparameterizable} by ensuring each rule length does not exceed $K$. In this way, 
the learnable parameters to determine $\Fcal$ are now becoming an unbounded weight matrix $W \in (R^+)^{H \times (|\Xcal|+M)}$, which can be learned by gradient-descent. Sepcifically, we leverage simulated annealing by tuning the temperature, which balances the approximation errors and issues caused by gradient vanishing in training.
\end{itemize}
In summary, we have performed the following series of approximations: 
{\begin{align}
 \underbrace{ \Fcal}_{\text{rule set}} \to  \underbrace{ A}_{\text{binary}} \to  \underbrace{ \tilde{A}}_{\text{selection probability}} \to  \underbrace{W}_{\text{unbounded weight}}
\end{align}}

and in the end we optimize $W$ end-to-end in a differentiable way to determine $\Fcal$. We provide a more detailed explanation as follows. 
\subsubsection*{I: Continuous Relaxation for Rule Learning}
We first formulate the rule learning as a variable selection problem and introduce a binary matrix $A=[a_{hj}]$ to encode the rule content (with columns indicating the logic variables and rows indicating the rules):
{\small\begin{equation} A=
\bordermatrix{%
       & X_1   & X_2     & \ldots & X_{|\Xcal|} &\xi_1 &\ldots &\xi_{M} \cr
f_1    & 0     & 1       & \ldots & 1 & 0 & \ldots & 1 \cr
\vdots & \vdots & \vdots & \ldots & \vdots & \vdots & \ldots & \vdots \cr
f_H    & 1      & 0       & \ldots & 1 & 1 & \ldots & 0   
}.
\label{eq:binary_matrix_A}
\end{equation}}

\begin{figure}[ht] 
\centering
\includegraphics[width=0.43\textwidth]{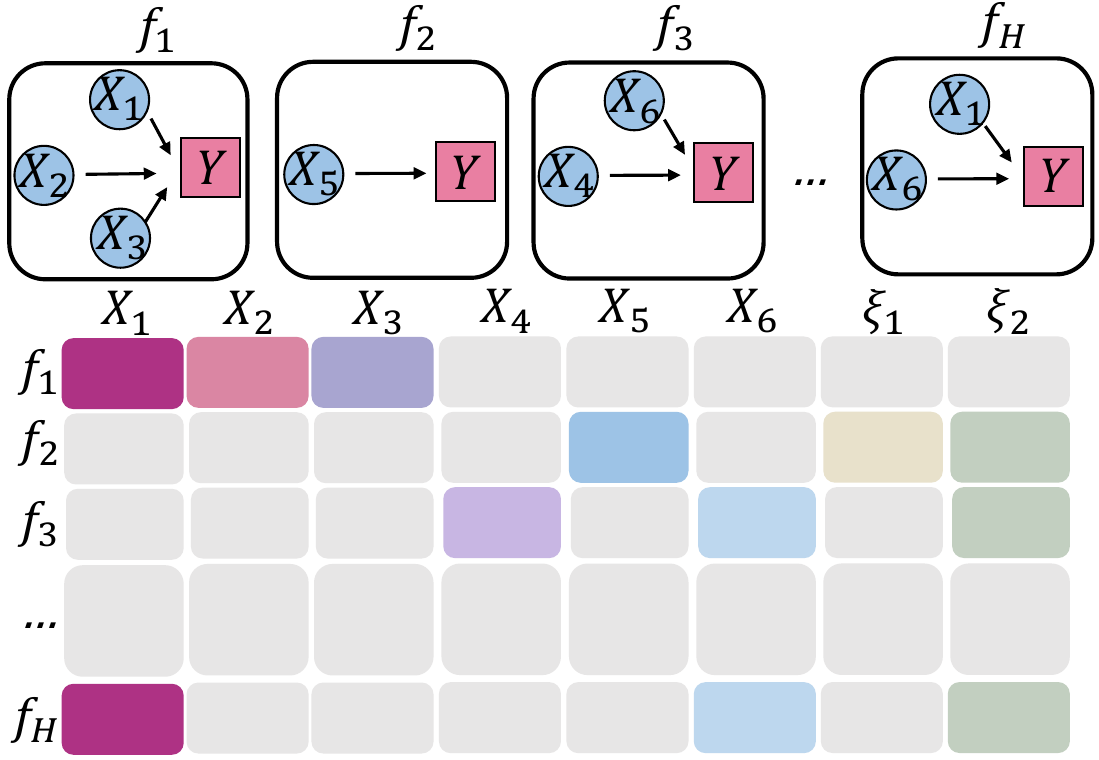}
\vspace{-10pt}
\caption{{\small Illustration of how to encode the rule content into a binary matrix $A$. In the top diagrams, the content for a total of $H$ rules is depicted. In the bottom diagram, the corresponding encoded binary matrix $A$ is demonstrated. The colorful areas in the matrix indicate a value of 1, while the grey areas indicate a value of 0. Each row in the matrix represents a rule, and each column corresponds to a predicate. $\xi_1$ and $\xi_2$ are dummy predicates, and $K$=3 in this example.}}
\label{rule_content}
\end{figure}

To accommodate for various rule lengths and meanwhile ensure that the rule length does not exceed $K$, we append $M$ (a hyperparameter) dummy columns after $X_{|\Xcal|}$, indicating {\it empty-predicates} to adjust for various rule lengths. We let $\sum_{j=1}^{|\Xcal|+M} a_{hj}=K$, and whenever $a_{hj}=1$ for $j >|\Xcal|$, it means that the rule length, determined by counting the valid predicates, will be less than $K$. A better illustration can be found in Fig.\ref{rule_content}. 

Learning such a binary $A$ with the constraints, $\sum_{j=1}^{|\Xcal|+M} a_{hj}=K$, $\forall h$, involves challenging discrete optimization. Instead, we will relax the problem to learning the selection probability matrix $\tilde{A}=[\tilde{a}_{hj}]$ with $0\leq \tilde{a}_{hj} \leq 1$. 
Similarly, we want to encode the temporal relations with the same idea. Given the selected predicates, we can further identify their temporal relations. For each selected paired $X_u$ and $X_v$, we introduce a learnable parameter $\boldsymbol{\alpha}$ to denote their corresponding probabilities
$$ \boldsymbol{\alpha} = [ \alpha_b, \alpha_e, \alpha_a, \alpha_n ]
$$
where $\alpha_b+\alpha_e+\alpha_a+\alpha_n=1$ and $0 \leq \alpha_b, \alpha_e, \alpha_a, \alpha_n \leq 1$. That is, $\boldsymbol{\alpha}$ belongs to a 3-$d$ simplex, indicating the probability that the temporal relation belongs to ``Before'', ``Equal'', ``After'', or ``None''. We represent the selection probabilities for all temporal relations in a rule as $\boldsymbol{\alpha_h}$, and these learned values determine the types of temporal relations.

\subsubsection*{II: Continuous Approximation of the Boolean Logic Features}
To learn the rule set in a differentiable way in the M-step, we propose to approximate the log-likelihood function as a deterministic function of $\tilde{A}$ and $[\boldsymbol{\alpha_h}]_h$. This is realized by approximating the boolean logic feature (as shown in Eq.(\ref{eq:feature})) and rewriting the intensity function as 
{\begin{align}
\lambda_{\text{soft}}^*(t_i \mid  \bm{z}_i) = b_0 + \sum_{h=1, \dots, H} \gamma_h z_{ih}\Kcal_{h}{ \Tcal}_{h}
\label{eq:intensity1}
\end{align}}

where $\Kcal_{h}$ and ${ \Tcal}_{h}$ are deterministic functions of $\tilde{A}$ and $[\boldsymbol{\alpha_h}]_h$, respectively.

For example, we choose to compute $\Kcal_{h}$ by Laplace kernel denoted as $\Kcal_{Lap}$, such as 
{\begin{align}
\Kcal_{h} = \Kcal_{Lap}\left\{ \sum_{j=1}^{|\Xcal|+M} X_{j}(t_j)\tilde{a}_{hj},K \right\}
\end{align}}
where we use $\Kcal_{Lap}$ with center $K$ to reparameterize the boolean feature and make it differentiable with $\tilde{A}$. The value of this Laplace kernel approaches 1 if $\sum_{j=1}^{|\Xcal|+M} X_{j}(t_j)\tilde{a}_{hj}=K$, which indicates each selected predicate should be grounded as true (for dummy predicates $\xi$, we just let each $\xi=1$). In this way, we use $\Kcal_{Lap}$ to approximate the boolean feature for the static predicate parts. 

Furthermore, we introduce ${ \Tcal}_{h}$ to provide a softened approximation for temporal relations. We soften the relation $R_j(t_u, t_v)$ by replacing it with 
{\begin{align}
& \text{softmax} \left\{\alpha_b R_b\left(t_u,t_v\right), \alpha_e R_e\left(t_u,t_v\right), \alpha_a R_a\left(t_u,t_v\right), \right.\nonumber \\
& \left.\alpha_n \left(1-\alpha_b R_b\left(t_u,t_v\right)-\alpha_e R_e\left(t_u,t_v\right)-\alpha_a R_a\left(t_u,t_v\right)\right)\right\}.
\end{align}}

For each rule, we consider the temporal relations between all possible pairs of selected predicates. If a rule involves multiple combinations of two distinct predicates, we employ the softmin operation on these $R_j$ values. Using softmin helps prevent excessive value reduction resulting from multiple multiplications. The softmin value for a rule $h$ is denoted as ${\Tcal}_h, h=1,\dots,H$. The combination of softmax and softmin operations ensures that the probability learning process for $[\boldsymbol{\alpha_h}]_h$ remains differentiable. 

Up to this stage, one may also choose to directly optimize $\tilde{A}$ and $[\boldsymbol{\alpha_h}]_h$ by maximizing the expected likelihood, which will be formulated as a constrained optimization problem. However, sampling the rules from the selection probability will encourage exploration and may help get rid of bad suboptimal solutions.
\subsubsection*{III: Differentiable Top-$K$ Subset Sampling}
Given the above description, we propose to generate rules given $\tilde{A}$ in the forward pass computation, which is formulated as a top-$K$ subset sampling process.  We will use a reparameterization trick for this subset sampling. 

Assume that the top-$K$ subset sampling can be parameterized by an unbounded weight matrix $W=[w_1; w_2; \dots; w_H] \in (R^+)^{H \times (|\Xcal|+M)}$, with each element non-negative~\citep{xie2019reparameterizable}. The top-$K$ relaxed subset sampling (using Gumbel noise) is summarized in Alg.\ref{Alg.1}. By injecting the Gumbel noise, the introduced key $\hat r_j$ is differentiable with respect to the input weight $w_h$. We can use a top-$K$ relaxation based on successive applications of the softmax function~\citep{plotz2018neural} to obtain the relaxed top-$K$ vector $\tilde{a}_h$, with the summation constraints automatically satisfied. Specifically, temperature $\tau>0$ and it returns a relaxed $K$-hot vector $\tilde{a}_h$, such that as $\tau \rightarrow 0$, RelaxedTopK $(\hat{\mathbf{r}}, K, \tau) \rightarrow \sum_{j=1}^K \operatorname{TopK}(\hat{\mathbf{r}}, K)[j]$.

\begin{algorithm}[ht]
\caption{Relaxed Subset Sampling} 
\label{Alg.1} 
\renewcommand{\algorithmicrequire}{

\textbf{Input:}

} 
\renewcommand{\algorithmicensure}{

\textbf{Output:}
 } 
\begin{algorithmic}
\REQUIRE Weight vector $ w_h =[w_{h1}, w_{h2}, \dots, w_{hn}]$, subset size $K$, temperature $\tau > 0$
\ENSURE Relaxed $K$-hot vector $\tilde{a}_h=[\tilde{a}_{h1},\dots,\tilde{a}_{hn}]$, where $\sum_{j=1}^{n} \tilde{a}_{hj}=K, 0 \leq \tilde{a}_{hj}\leq 1$
\STATE $\mathbf{\hat r} \gets [\quad]$
\FOR {$j \gets 1$ to $n$} 
\STATE $u_j \gets \text{Uniform}(0,1)$
\STATE $\hat r_j \gets -\log(-\log(u_j)) + \log(w_{hj})$
\STATE \textbf{$\hat r$}.append($\hat r_j$)
\ENDFOR
\STATE $\tilde{a}_h \gets \text{RelaxedTopK}(\mathbf{\hat r},K,\tau)$
\end{algorithmic}
\end{algorithm}
Overall, to optimize the sampling matrix $W$, we have
{\begin{equation}
 \textbf{M-step:}\quad   \frac{\partial \ell(\theta_0,W)}{\partial w_h} = \frac{\ell(\theta_0, W)}{\partial \tilde{a}_{h}} \frac{\partial \tilde{a}_h}{\partial \hat r_h}\frac{\partial \hat r_h}{\partial w_h} ,
\end{equation}}

where $\frac{\ell(\theta_0, W)}{\partial \tilde{a}_{h}}$ is computed based on the approximation as discussed in (II), and the latter two partial gradients are obtained based on the differentiable top-K sampling. The parameter $W$ is updated with stochastic gradient descent.

With this reparameterization trick, the level of discreteness of $A$ can be controlled by the temperature parameter $\tau$. Thus, in our algorithm, an almost discrete matrix $A$ is used to select rules, and the weight matrix $W$ is updated in every iteration to sample $A$. During the learning process, we employed a cyclical annealing scheme to fine-tune the temperature parameter $\tau$. The choice of $\tau$ balances the trade-off between approximation errors and gradient vanishing issues. Each time when we iterate all data, $\tau$ anneals from a large value to a small non-zero one. 

Rule content parameter $W$ is optimized first until convergences by ignoring the temporal relations. After $W$ converges, we will proceed to optimize $[\boldsymbol{\alpha_h}]_h$ until convergence. The number of rules $H$ and rule length range $K$ can be tuned via cross-validation or predefined with domain knowledge.
\subsubsection{Optimization w.r.t. \texorpdfstring{$\theta_0$}{theta\_0}}
{\bf M-step:} We optimize other model parameters by taking gradient of $\ell(\theta_0, W)$ w.r.t. $[ b_0,[r_h]_{h \in [1,\dots,H]},\bm{\pi}]$, and set these gradients to 0. We need to make sure that these model parameters are non-negative. This can be guaranteed by projected gradient descent or using the change-of-variable trick (e.g., parameterize $b_0$ as $\exp(b_0)$). 

Specifically, solving $\bm{\pi}$ in the M-step has a closed-form solution. Given the constraints $\sum_{h=0}^H \pi_{h}=1$ we have 
{\begin{equation}
 \textbf{M-step:}\quad\pi_{h}= \frac{n_h}{n}, \quad  n_h =\sum_{i=1}^n Q^{new}(z_{ih}=1). \end{equation}}

\section{Experiments}
\subsection{Synthetic Data Experiments} 
We assess our model's learning accuracy in terms of rule discovery, rule weight estimation, and rule prior distribution learning across various scenarios. We also evaluate the inference accuracy by identifying the most likely causal rules for each event. 

\textbf{I: Experiment Setup}\\ In our empirical analysis, we systematically explore various aspects to evaluate the scalability and performance of our method: ({\it i}) {\bf Diverse Ground Truth Rules:} We vary the number of ground truth rules, ranging from 1 to 4, to examine the model's performance across different complexity levels. ({\it ii}) {\bf Variation in Predicate Set:} We consider five scenarios with different sizes of predicate sets, including 10, 15, 20, 25, and 30, to assess the scalability of our method. A larger logic variable set represents a more challenging problem. ({\it iii}) {\bf Variation in Data Size:} We conduct experiments with varying data sizes, including sequences of 5000, 10,000, 20,000, 30,000, and 40,000 instances, to investigate the EM algorithm's scalability under different data scales.

It is important to note that only a small subset of the predicates is included in the ground truth rules. In other words, the majority of candidate logic variables are irrelevant. We aim to demonstrate that our method excels in pinpointing the most critical logic variables during the rule-learning process. We present the primary results based on experiments with 20,000 sequences for each scenario. For details and comprehensive findings, please refer to Appendix Sec.\ref{sec:all_rules_synthetic_data}.

\begin{figure}[ht]
\centering
\includegraphics[width=0.5\textwidth]{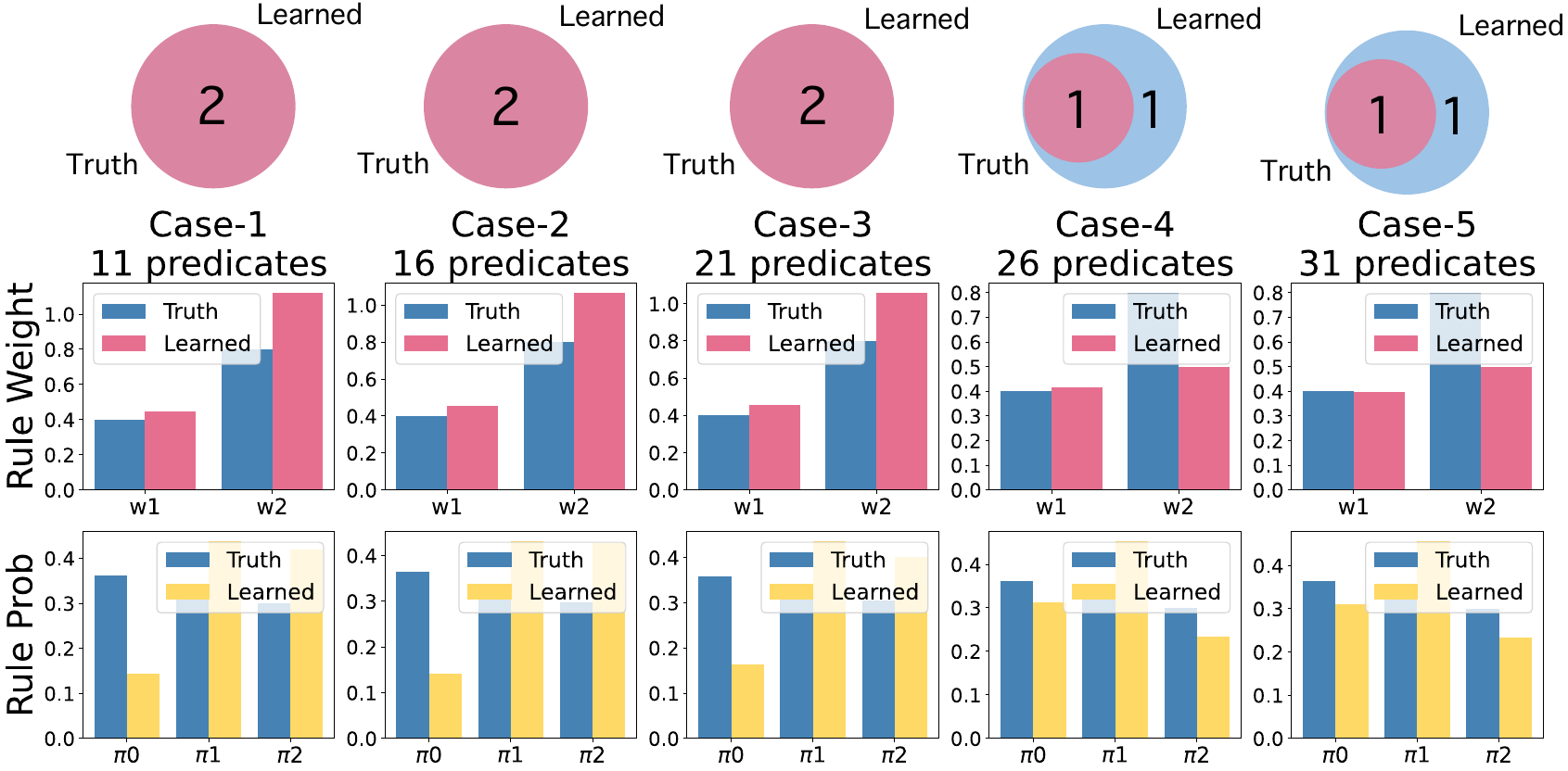}
\caption{{\small Our proposed model's performance is evaluated across all five scenarios for group 2, with two ground truth rules. We evaluate our model's performance in terms of rule discovery, rule weight learning, and rule prior distribution learning. The color ``blue'' indicates ground truth rules, weights, and prior distributions, whereas the colors ``red'' and ``yellow'' indicate the learning results.}}
\label{synthetic_results_2rules}
\end{figure}

\begin{figure}[ht]
\centering
\includegraphics[width=0.5\textwidth]{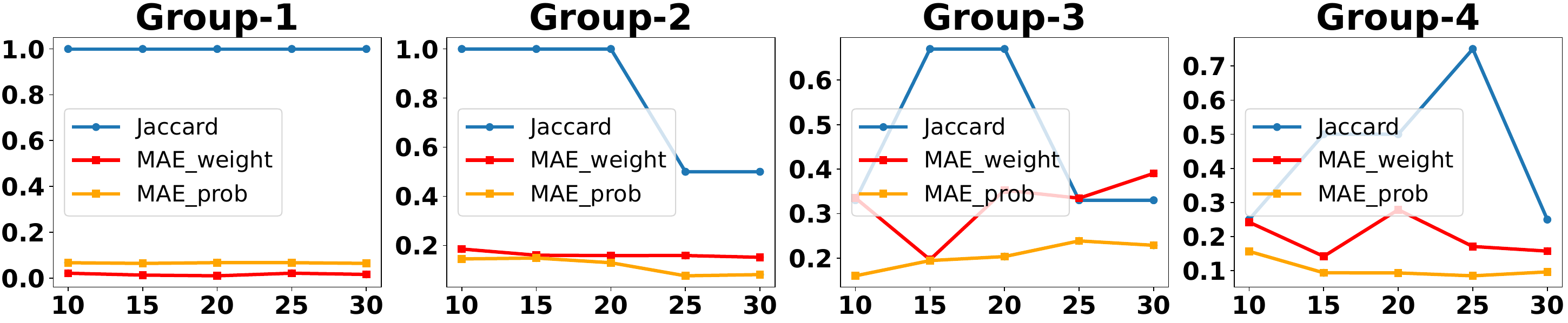}
\vspace{-5pt}
\caption{{\small Jaccard similarity score and MAE of rule weights and rule prior probabilities for all 4 groups. The X-axis indicates predicate library size and the Y-axis indicates the value of Jaccard similarity and MAE.}}
\label{line_plot}
\end{figure}

\textbf{II: Accuracy} \\ We evaluate the accuracy of our method in terms of rule set discovery and TPP model parameter learning. In particular, we assess its accuracy in increasingly challenging scenarios by gradually enhancing the number of irrelevant logic variables in the predicate set.

In Fig.\ref{synthetic_results_2rules}, we present the results for group-2, while the complete results are available in the Appendix Sec.\ref{sec:all_rules_synthetic_data}. The top row of the figure employs Venn diagrams to visually depict the agreement between the actual rule set and the learned rule set, quantified through the Jaccard similarity score (the intersection area divided by the union area). A rule is deemed correct when both its included predicates and temporal relations are accurately learned. Our proposed model consistently identifies almost all the ground truth rules across all groups.

The middle row of figures compares the true rule weights with the learned rule weights, and the bottom row of diagrams compares the true rule prior distributions with the learned rule distributions. Once the ground truth rules are accurately learned, the learned prior rule distributions closely align with the true values, resulting in low mean absolute errors (MAEs). The MAE is calculated treating weights and priors as vectors for each case.

In addition, we also compared our model with several classic unsupervised rule mining methods, including Apriori~\citep{agrawal1994fast}, NEclatcloesed~\citep{aryabarzan2021neclatclosed}, CM-SPADE~\citep{fournier2014fast} and VGEN~\citep{fournier2014vgen}. These baseline methods identify rules using the frequency thresholds, resulting in coarse rule discovery. From the experiment results, these approaches lead to the extraction of a vast number of noisy and erroneous rules, while the few accurate rules tend to remain hidden within the extensive rule set. A complete results and discussion can be found in Appendix Sec.\ref{sec:baselines} and Sec.\ref{sec:compare_with_other_baselines}.

\textbf{III: Scalability}\\ 
Fig.\ref{line_plot} illustrates the Jaccard similarity scores and MAE for all groups as the number of redundant predicates increases. Across all four groups, we observe a decrease in the Jaccard similarity score and a slight increase in MAE with the growing number of predicates. However, it's important to note that our model maintains stability and reliability when the size of the predicate set is appropriate (i.e., around 20-30) and adequate event sequences are available.

Furthermore, our approach exhibits promising results in identifying the latent causal rules for each event. For each event, we denote the rule inferred by our learned rule set as $f$ and the true rule as $f^*$. Each rule is a binary vector comprised of two parts, one of which tracks rule content, as shown in Eq.\ref{eq:binary_matrix_A}, while the other records temporal relation. For a pair of predicates, we compute the cosine similarity score $\frac{f^{\top}f^{*}}{\left \| f \right \|\left \| f^{*} \right \|} \in \left [ 0, 1 \right ]$ for each event. We then take the average over all sequences and report the results in Tab.\ref{tab:posterior accuracy_synthetic_data}. Notably, in scenarios where almost all ground truth rules were successfully learned, the similarity score is exceptionally high. However, in more challenging tasks, such as those in Group-3 and Group-4, where a larger number of ground truth rules are involved, the similarity score of the rule index posterior experiences a slight decrease. 

\begin{figure}[ht]
\centering
\includegraphics[width=0.4\textwidth]{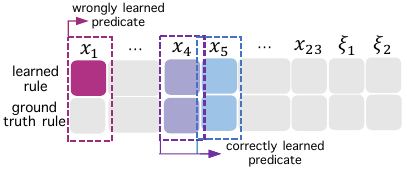}
\vspace{-10pt}
\caption{{\small One example of inaccurately uncovered rule for Group-3 (3 ground truth rules) Case-4 (23 to-be-searched predicates). Ground truth rule: $Y \leftarrow x_4 \land x_5 \land ( x_4\,\text{Before}\, x_5)$. Learned rule: $Y \leftarrow x_1 \land x_4 \land x_5 \land(\ x_4 \,\text{Before} \ x_5)$. We see that only one more predicate was wrongly excluded.}}
\label{wrongly_learned_rule_explaination}
\end{figure}

We need to address that our discovered rules do not deviate significantly from the true rules even if they don't match exactly. Fig.\ref{wrongly_learned_rule_explaination} provides an example of a learned wrong rule from Group-3, Case-4. In this instance, the discovered inaccurate rule includes only one additional redundant predicate. Similar patterns are observed in other cases as well.

\vspace{5pt}
\begin{table}[ht]
\centering
\resizebox{\columnwidth}{!}{%
\begin{tabular}{c|cccc} 
\toprule
\diagbox{Case}{Group} & Group-1 & Group-2 & Group-3 & Group-4 \\ 
\hline
      1 (10preds) & 0.8132 & 0.7890 & 0.5033 & 0.4879\\
      2 (15preds) & 0.7916 & 0.7835 & 0.4903 & 0.4900\\
      3 (20preds) & 0.7992 & 0.7708 & 0.4889 & 0.4738\\
      4 (25preds) & 0.7983 & 0.7011 & 0.4799 & 0.4729\\
      5 (30preds) & 0.8007 & 0.6874 & 0.4329 & 0.4295\\
      \bottomrule
\end{tabular}
}
\caption{{\small Rule index average inference cosine similarity score on synthetic data for all groups and all cases. For each case we repeated the experiment for 20 times to reduce the randomness.}}
\label{tab:posterior accuracy_synthetic_data}
 \end{table}
 
\textbf{IV: Event Prediction}\\
We compared our model with several state-of-the-art baselines in terms of event predictions, using MAE as the evaluation metric for event time prediction. A complete description of these baseline methods can be found in Appendix Sec.\ref{sec:baselines}. As shown in Tab.\ref{tab:prediction_event_time_synthetic_data}, our model outperforms all baselines, particularly in complex scenarios like Case-5 with 30 predicates across all groups. In other cases, the MAE varies depending on rule learning outcomes, but it consistently matches or surpasses the baselines.
\begin{table}[ht] 
\centering
\resizebox{\columnwidth}{!}{%
\begin{tabular}{c|cccc} 
\toprule
\diagbox{Method}{Group} & Group-1 & Group-2 & Group-3 & Group-4 \\ 
\hline
      THP~\citep{zuo2020transformer} & 2.2155 & 2.0089 & 2.2364 & 2.4121 \\
      RMTPP~\citep{du2016recurrent} & 2.2354  & 2.1843 & 2.3147 & 2.4523 \\
      ERPP~\citep{xiao2017modeling} & 2.3532 & 2.0175 & 2.3564 & 2.4264 \\
      GCH~\citep{xu2016learning} & 2.5233  & 2.3754 & 2.4655  & 2.6427\\
      LG-NPP~\citep{zhang2021learning} & 2.4145  & 2.4732 & 2.6564 & 2.5612 \\
      GM-NLF~\citep{eichler2017graphical} & 2.3147  & 2.3622 & 2.4522 & 2.4362 \\
      TELLER~\citep{li2021explaining} & 2.2281 & 2.4812 & 2.5121  & 2.4030 \\
      CLNN~\citep{yan2023weighted} & 2.3858 & 2.5165 & 2.4972  & 2.5096 \\
     \bf{OURS*} & \bf{2.2075}  & \bf{1.8586} & \bf{2.1957} & \bf{2.3559} \\
      \bottomrule
\end{tabular}
}
\caption{{\small Event time prediction MAE on synthetic data for complex Case-5 (30 predicates) of all 4 groups.}}
\label{tab:prediction_event_time_synthetic_data}
 \end{table}

\subsection{Healthcare Data Experiments}
MIMIC-IV\footnote{\url{https://mimic.mit.edu/}} is a dataset of electronic health records for patients who have been admitted to the ICU~\citep{johnson2023mimic}. In our study, we focused on 4074 patients diagnosed with sepsis \citep{saria2018individualized}, as sepsis is a leading cause of mortality in ICU, particularly when it progresses to septic shock. Septic shocks are critical medical emergencies, and timely recognition and treatment are crucial for improving survival rates. The objective of this experiment is to identify logic rules associated with septic shocks for both the general population and specific patients. These rules could serve as potential early alarms when abnormal indicators are detected, aiding in timely intervention.

\begin{table}[ht] 
\centering
{\small \begin{tabular}{|c|c|}
\hline Weight & Rule \\
\hline \hline 0.566 & \begin{tabular}{c} 
Rule 1: LowUrine $\leftarrow$ \\
Arterial Blood \\
Pressure Diastolic
\end{tabular} \\
\hline 0.467 & \begin{tabular}{c} 
Rule 2: LowUrine $\leftarrow$ \\
Ionized Calcium
\end{tabular} \\
\hline 0.527 & \begin{tabular}{c} 
Rule 3: LowUrine $\leftarrow$ \\
Arterial CO2 Pressure
\end{tabular} \\
\hline 0.466 & \begin{tabular}{c} 
Rule 4: LowUrine $\leftarrow$ \\
Venous 02 Pressure
\end{tabular} \\
\hline 0.464 & \begin{tabular}{c} 
Rule 5: LowUrine $\leftarrow$ \\
Respiratory Rate $\wedge$ Hemoglobin
\end{tabular} \\
\hline 0.664 & \begin{tabular}{c} 
Rule 6: LowUrine $\leftarrow$ Heart Rate $\wedge$ BUN $\wedge$ \\
WBC $\wedge$ (Heart Rate  Before BUN)\\
$\wedge$ (BUN Before WBC)
\end{tabular} \\
\hline
\end{tabular}}
\caption{{\small Some examples of our discovered temporal logic rules in MIMIC-IV.}}
\label{tab:mimic_low_urine}
 \end{table}

\textbf{Routine Vital Signs and Lab Values} 
A total of 29 variables associated with sepsis were extracted, including routine vital signs and laboratory values, as suggested and utilized by \cite{komorowski2018artificial}. Detailed illustration can be found in Appendix Sec.\ref{sec:variables_describe}. We recorded the time points at which these variables first exhibited abnormal values within the 48 hours preceding the onset of abnormal urine output (i.e., our target event).

\textbf{Discovered Logic Rules} Real-time urine output was utilized as the health state indicator, as low urine output directly indicates a deteriorated circulatory system and serves as a warning sign for septic shock. Due to the frequent fluctuations in urine output within the ICU setting, only instances where urine output becomes abnormal after maintaining a normal level for at least 48 hours were considered valid target events (i.e., meaningful to predict and explain).

In Tab.\ref{tab:mimic_low_urine}, we present a portion of the discovered explanatory logic rules and their corresponding learned weights. These displayed rules have been justified by doctors or supported by relevant literature, demonstrating their clinical significance. 




Our methodology incorporated invaluable viewpoints from experts who thoroughly reviewed our discovered rules. Their feedback consistently validated the soundness and relevance of our findings. Additionally, we bolstered our discovered rules by referencing supporting evidence from reputable medical sources, with detailed discussions and explanations found in Appendix Sec.\ref{sec:doctor_verification}.

We also compared our model with the post-hoc method~\citep{crabbe2021explaining} in terms of interpretability, with the corresponding results in Appendix Sec.\ref{sec:post-hoc}. While the post-hoc methods are easier to implement, our model surpasses them in terms of accuracy and interpretability. Moreover, our model remains reliable even in small data scenarios, which is challenging for post-hoc methods relying on black-box models to achieve.

\begin{figure}[ht]
\centering
\includegraphics[width=0.5\textwidth]{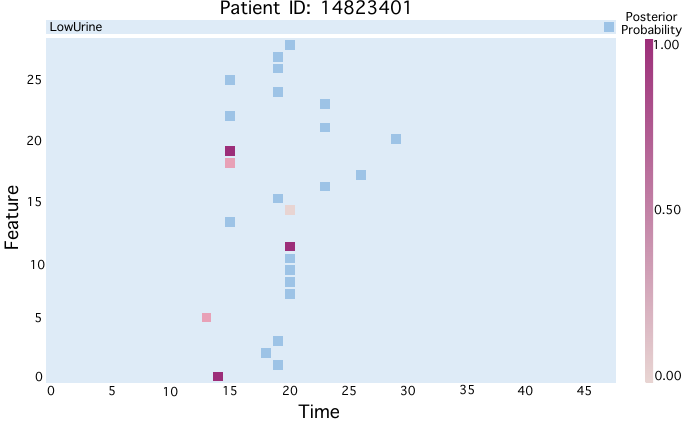}
\vspace{-20pt}
\caption{{\small Rule identification results for patient 14823401. X-axis: time ranges from 0 to 48h. Y-axis: feature ID (29 in total). Low urine output occurred at 48h. Deep blue blocks record the occurrence time of a routine vital sign or abnormal lab value. Identified rule with largest posterior probability (colored in purple), Rule 6: LowUrine $\gets$ \texttt{Heart Rate} $\land$ \texttt{BUN} $\land$ \texttt{WBC}, \texttt{Heart Rate} \texttt{Before} \texttt{BUN}, \texttt{BUN} \texttt{Before} \texttt{WBC}. Other rules with relatively large posterior probability are colored in pink and off white. The darker the color, the greater the posterior probability.}}
\label{clustering_visulization}
\end{figure}

\textbf{Logic Rule Identification for Each Event} Our method can identify the most likely causal rule specific to individual patients. For instance, let's consider the patient with ID 14823401 as an example, and the results are visualized in Fig.\ref{clustering_visulization}. Based on the analysis, the rule with the highest posterior probability that the patient is likely to adhere to is Rule 6: LowUrine $\gets$ \texttt{Heart Rate} $\land$ \texttt{BUN} $\land$ \texttt{WBC}, indicating that this patient requires closer monitoring of Heart Rate, BUN, and WBC levels exceeding the normal range in their daily life, which likely serves as an early warning sign. This example sufficiently illustrates the effectiveness of our method at the individual patient level, showcasing its potential for individual healthcare delivery and treatment design. 
\textbf{Event Prediction} We used the same baselines as in the synthetic data experiments, with the MAE serving as the evaluation metric for predicting the occurrence time of ``Low Urine'' events. Our model's performance is presented in Tab.\ref{tab:prediction_event_time_mimic_data}, from which one can see that our model outperforms all other baselines.

\begin{table}[ht]
\centering
{\small \begin{tabular}{c|cc} 
\toprule
Method & LowUrine \\ 
\hline
      THP &  2.4234 \\
      RMTPP & 2.4643 \\
      ERPP & 2.6122 \\
      GCH & 2.5367 \\
      LG-NPP & 2.5672 \\
      GM-NLF & 2.6925 \\
      TELLER & 2.4401 \\
      CLNN &  2.4371 \\
     \bf{OURS*} & \bf{2.3675} \\
      \bottomrule
\end{tabular}}
\caption{{\small Event time prediction MAE (unit/hour) on healthcare data.}}
\label{tab:prediction_event_time_mimic_data}
 \end{table}

\section{Conclusion}
\label{sec:conclusion}
In this paper, we leverage the temporal point process models to effectively discover latent logic rules to predict and explain abnormal events. We design an EM algorithm that can jointly learn model parameters, discover a rule set, and infer the most likely rule factor for each event. We especially can learn the rule set in a differentiable way. Our method has exhibited strong performance, as demonstrated through synthetic and real datasets.


\subsubsection*{Acknowledgements}
›Shuang Li’s research was in part supported by the NSFC under grant No. 62206236, Shenzhen Science and Technology Program under grant No. JCYJ20210324120011032,  National Science and Technology Major Project under grant No. 2022ZD0116004, Shenzhen Key Lab of Cross-Modal Cognitive Computing under grant No. ZDSYS20230626091302006, and Guangdong Key Lab of Mathematical Foundations for Artificial Intelligence.
\bibliography{reference.bib}

\newpage
\onecolumn

\clearpage
\appendix
\setcounter{section}{0}
\section*{APPENDIX}

\renewcommand\thesection{\Alph{section}}

In the following, we will provide supplementary materials to better illustrate our methods and experiments.

\begin{itemize} 

\item Section \ref{sec:all_rules_synthetic_data} provides detailed analysis for synthetic data experiments, including experiment setup, implementation details, results analysis, and model scalability analysis.

\item Section \ref{sec:computing_infrastructure} provides the information of computing infrastructure for both synthetic data experiments and real-world data experiments.

\item Section \ref{sec:baselines} introduces the baseline methods we considered in this paper.

\item Section \ref{sec:post-hoc} provides the comparison of our method with a SOTA post-hoc method.

\item Section~\ref{sec:compare_with_other_baselines} provides the detailed comparison results compared with other rule mining baselines.

\item Section \ref{sec:variables_describe} describes the details of MIMIC-IV dataset preprocessing and risk factors extracting.

\item Section \ref{sec:doctor_verification} provides doctor verification and medical references towards our experiment results using MIMIC-IV datasets.

\end{itemize}

\section{Detailed Analysis for Synthetic Data Experiments}
\label{sec:all_rules_synthetic_data}

\textbf{Setup} In our empirical analysis, we systematically explore various aspects to evaluate the scalability and performance of our method: \textbf{(i) Diverse Ground Truth Rules:} We vary the number of ground truth rules, ranging from 1 to 4 (4 groups in total), to examine the model’s performance across different complexity levels. \textbf{(ii) Variation in Predicate Set:} We consider five cases with different size of predicate set, including 10, 15, 20, 25, and 30, to assess the scalability of our method. A larger logic variable set represents a more challenging problem. \textbf{(iii) Variation in Data Size:} We conduct experiments with varying data sizes, including sequences of 5000, 10,000, 20,000, 30,000, and 40,000 instances, to investigate our proposed EM algorithm’s scalability under different data scales. The source code for the implementation can be found at github\footnote{https://github.com/000kylie000/EventSequenceClustering} for further examination and review.

\newpage

\begin{figure}[ht]
\centering
\includegraphics[width=0.6\textwidth]{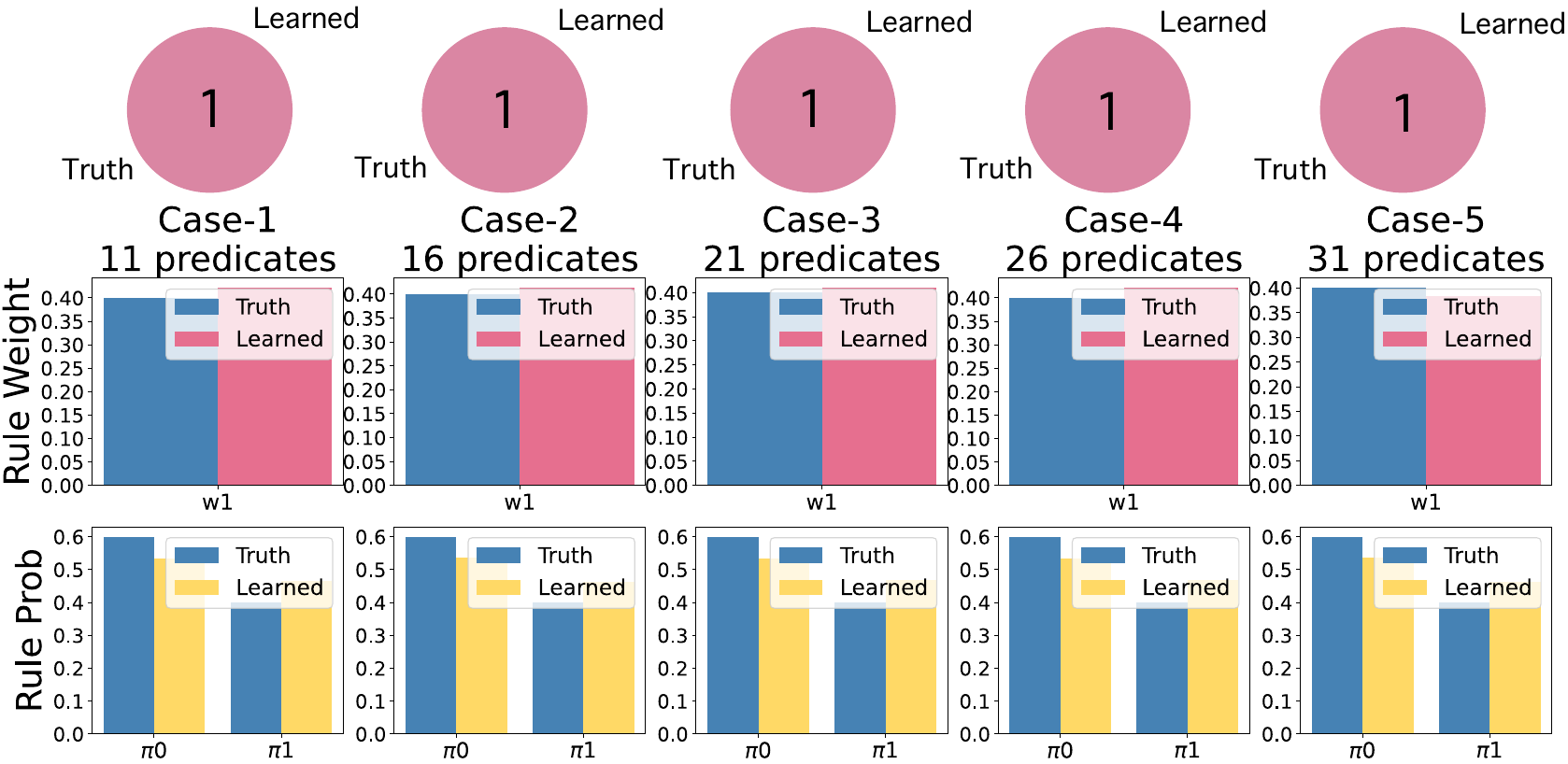}
\caption{Rule discovery ability, rule weight learning accuracy, and the accuracy of learning the probability of a rule appearing in the population of our proposed model on all 5 cases for group-1 (1 ground truth rules) using 20000 sequences. Blue one indicates ground truth rule/ground truth rule weight/ground truth probability of a rule appearing in the population, and red/yellow one indicates learned rule weight/learned probability of a rule appearing in the population.}
\label{synthetic_results_1rules}
\end{figure}

\begin{figure}[ht]
\centering
\includegraphics[width=0.6\textwidth]{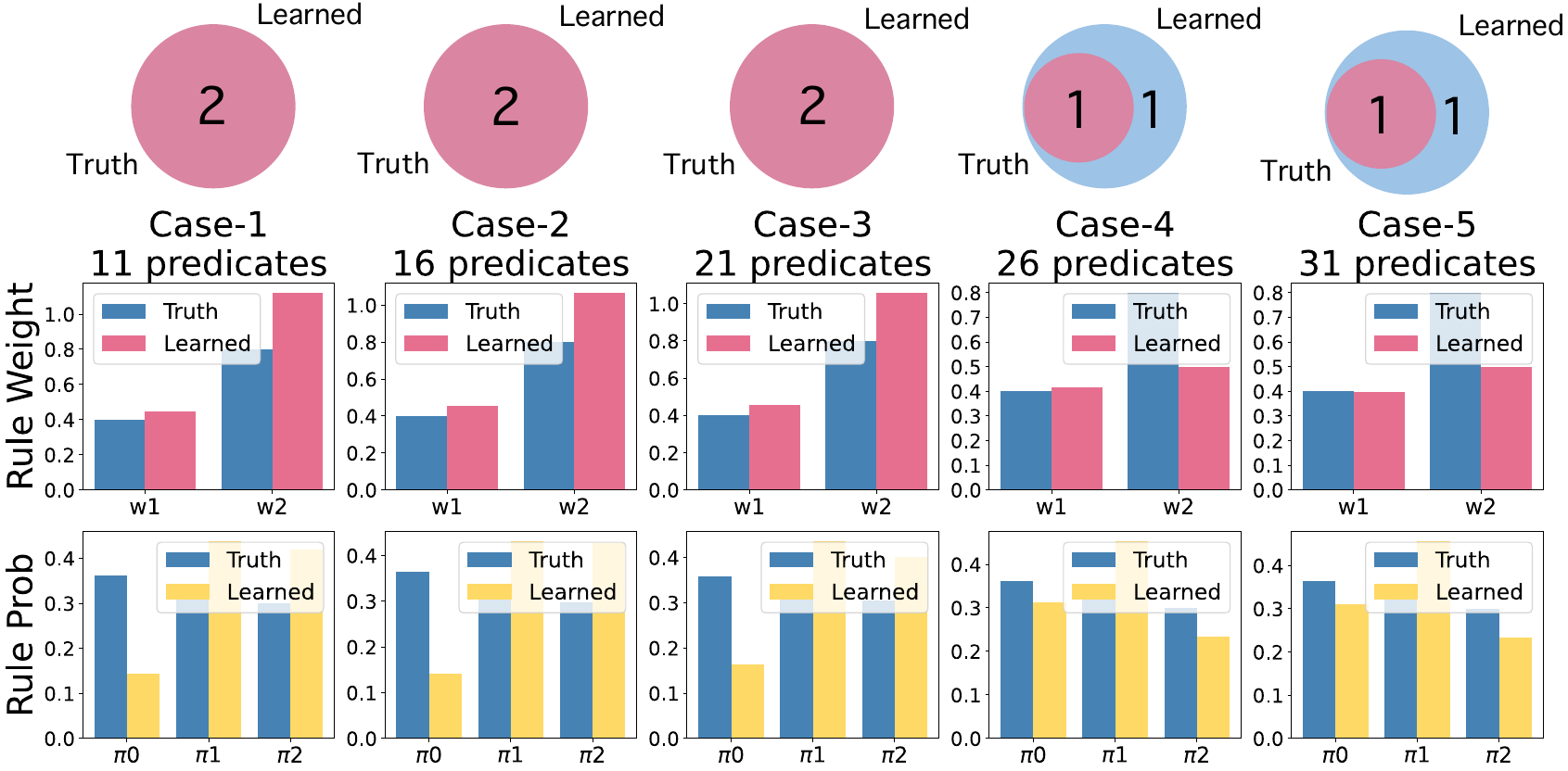}
\caption{Learning results for group-2 (2 ground truth rules).}
\label{synthetic_results_2rules_appendix}
\end{figure}

\begin{figure}[ht]
\centering
\includegraphics[width=0.6\textwidth]{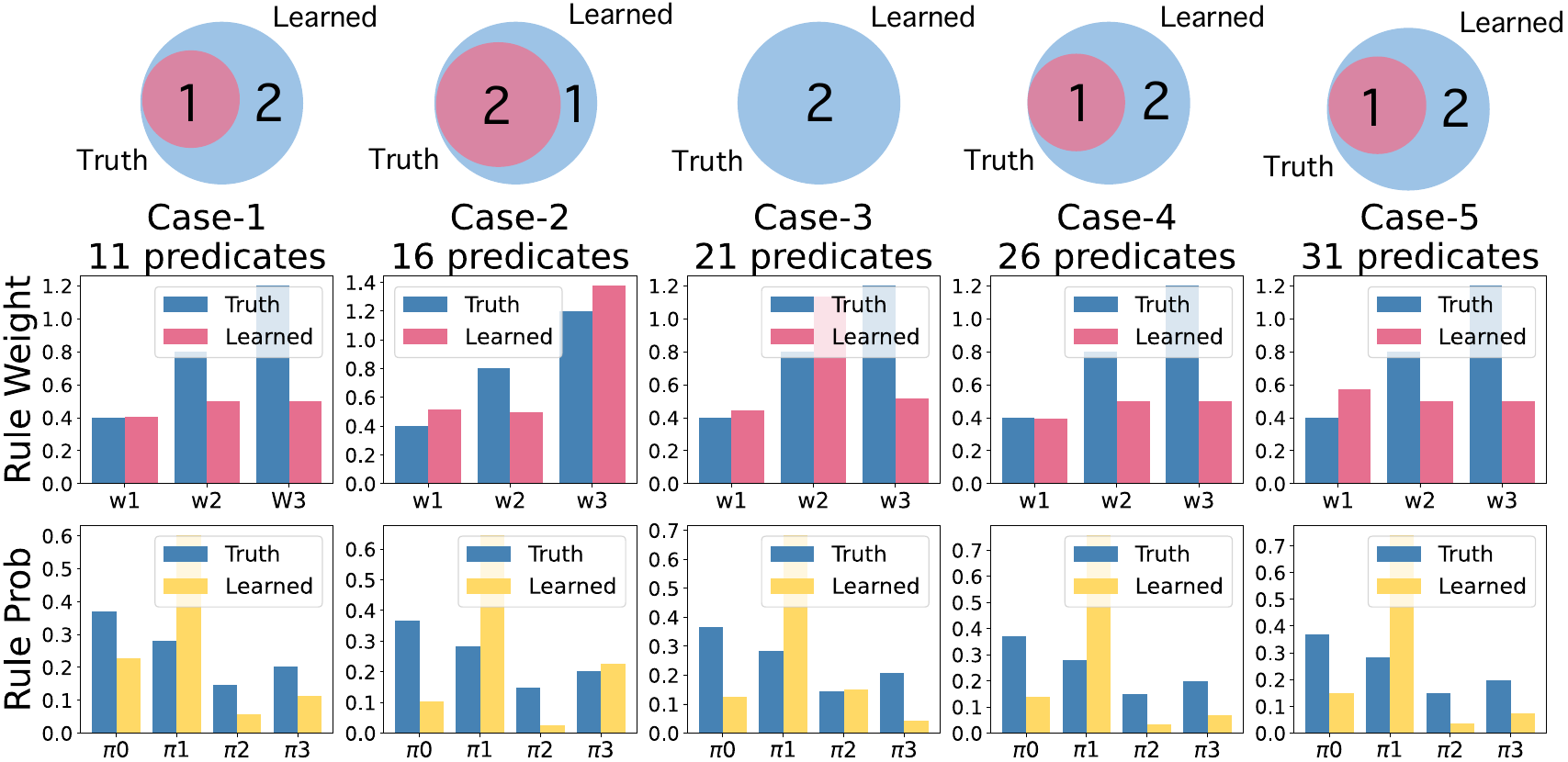}
\caption{Learning results for group-3 (3 ground truth rules).}
\label{synthetic_results_3rules}
\end{figure}

\begin{figure}[ht]
\centering
\includegraphics[width=0.6\textwidth]{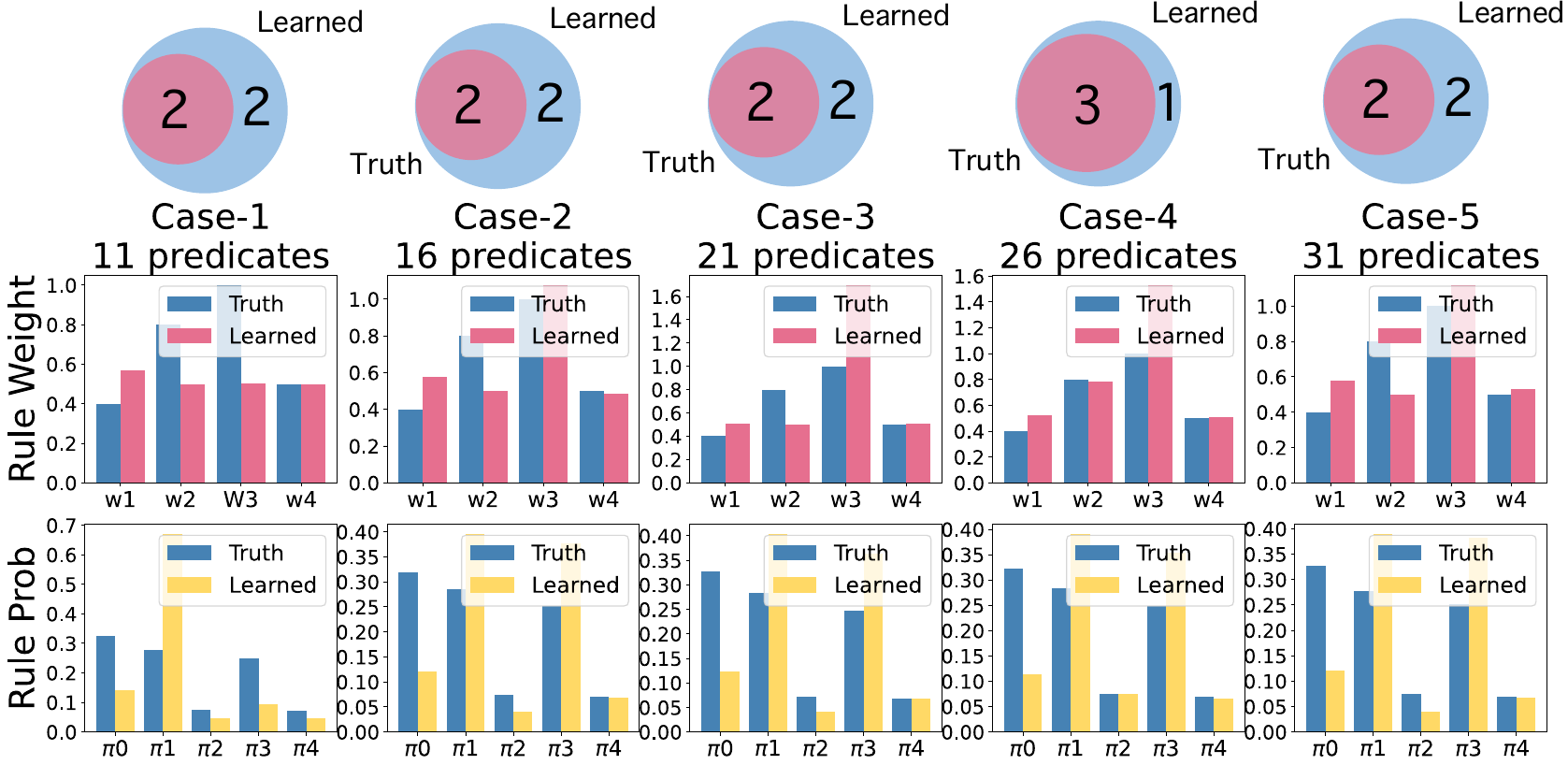}
\caption{Learning results for group-4 (4 ground truth rules).}
\label{synthetic_results_4rules}
\end{figure}

\begin{figure}[ht]
\centering
\includegraphics[width=0.8\textwidth]{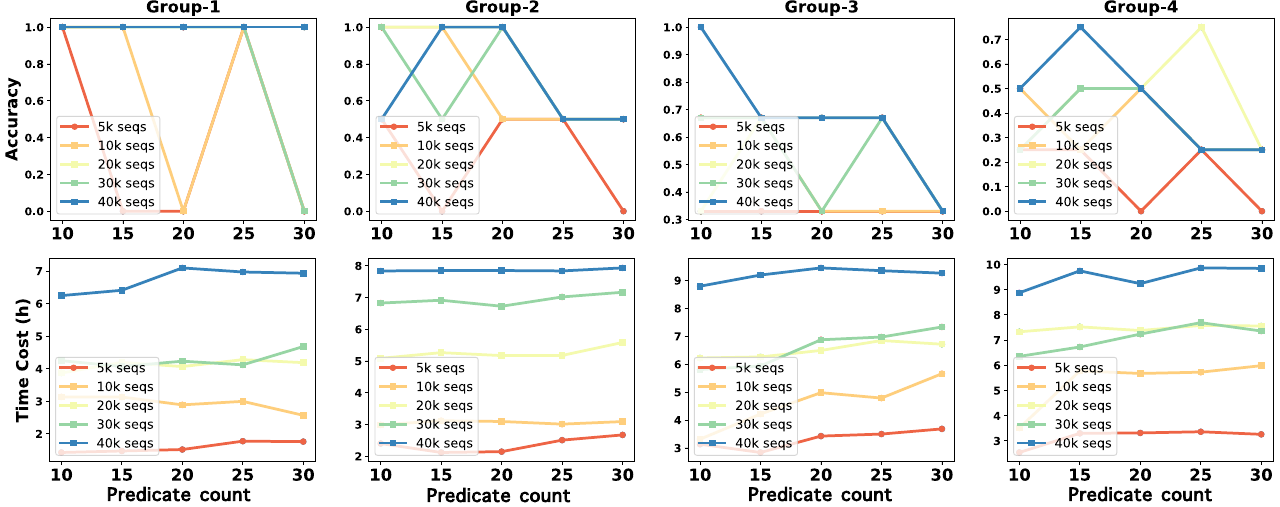}
\caption{Top: rule learning accuracy for each case. Bottom: time consumed for each case. }
\label{accuracy_and_time}
\end{figure}

We listed all the ground truth rules and discovered rules for all groups in Tab.\ref{appendix_tab:all_groups} using 20000 sequences for each case. In Fig.\ref{synthetic_results_1rules}-\ref{synthetic_results_4rules}, we report the discovery accuracy of the rule set , the learning accuracy of the rule weight parameters and rule prior  (i.e., the appearance prior probability of each rule in population) distribution parameters, using 20000 sequences for each case. In Fig.\ref{accuracy_and_time}, we report how the rule learning accuracy and the computation time change over different predicate sets and sample size.

\textbf{Implementation Details} 
The E-step, which refers to inference, has a closed-form.The M-step is more involved. In M-step, we learn model parameters, including the predicate assignment to rules, the temporal relationships among the selected predicates, the impact weight of each rule, and the prior distribution of the rules, using coordinate descent. In other words, we optimize one component of the model parameters by holding the others.  

The most critical aspect of this process involves selecting the predicates and determining the temporal relationships within a rule. To achieve a stable optimization, we set a small learning rate, specifically 0.0001, to prevent rapid changes in predicate selection. Additionally, to ensure comprehensive exploration of the rule set and fine-tuning of predicate weights, we cyclically anneal the temperature parameter, denoted as $\tau$ in Alg.~\ref{Alg.1}, during each iteration. In other words, as we iterate through the data, $\tau$ transitions from a large value to a small, non-zero temperature.
As for the learning rate pertaining to temporal relations, we set it slightly higher at 0.0035 to facilitate the rapid convergence of these relations.

\textbf{Results Analysis} 

Across all four groups of 20000 sequences, we observe that an increase in the number of redundant predicates within the same group leads to a slight decrease in rule learning accuracy, as well as reduced accuracy in rule weights and probabilities. In different groups, the presence of more underlying ground truth logic rules poses a greater challenge in accurately learning these rules in synthetic datasets. Therefore, when the number of ground truth logic rules increases while the number of redundant predicates remains constant, both rule learning accuracy and the accuracy of rule weights and probabilities decrease. However, despite the decline in experimental performance that occurs when the task is challenging, our model remains stable and reliable when the size of the predicate set is appropriate (i.e., around 20-30) and there are sufficient event sequences available.

It is crucial to emphasize that our reported final model parameter accuracy relies on the rule set discovery accuracy. If our algorithm learns an incorrect logic rule, we will compute the accuracy of the corresponding rule prior distribution and rule weight parameter as 0, which is very harsh.

In fact, our mined imprecise rules didn’t deviate significantly from the ground truth rules. Illustrated in Tab.\ref{appendix_tab:all_groups}, the discovered inaccurate rules only add or miss a small number of redundant predicates, with the majority parts of the true rule content successfully learned.

\begin{table}[ht]
  \centering
  	
\begin{tabular}{|c|c|c|c|c|}

\hline
 & Group-1 & Group-2 & Group-3 & Group-4\\ \hline
\textcolor{blue}{Ground Truth} & 
\tabincell{l}{
\textcolor{blue}{*} \texttt{$Y$}\ $\gets$\ \texttt{$X_1$}\ $\land$\ \texttt{$X_2$} \\ 
$\land$\ \texttt{$X_3$} \\
$\land$\ \texttt{$X_1$} \texttt{Before} \texttt{$X_2$} 
} &  
\tabincell{l}{
\textcolor{blue}{*} \texttt{$Y$}\ $\gets$\ \texttt{$X_1$}\ $\land$\ \texttt{$X_2$} \\ 
$\land$\ \texttt{$X_3$} \\
$\land$\ \texttt{$X_1$}\ \texttt{Before}\ \texttt{$X_2$} \\
\textcolor{blue}{*} \texttt{$Y$}\ $\gets$\ \texttt{$X_4$}\ $\land$\ \texttt{$X_5$} \\$\land$\ \texttt{$X_4$}\ \texttt{After}\ \texttt{$X_5$} 
} &  
\tabincell{l}{
\textcolor{blue}{*} \texttt{$Y$}\ $\gets$\ \texttt{$X_1$}\ $\land$\ \texttt{$X_2$} \\ 
$\land$\ \texttt{$X_3$}  \\

\textcolor{blue}{*} \texttt{$Y$}\ $\gets$\ \texttt{$X_4$}\ $\land$\ \texttt{$X_5$} \\
$\land$\ \texttt{$X_4$} \texttt{Before} \texttt{$X_5$}\\
\textcolor{blue}{*} \texttt{$Y$}\ $\gets$\ \texttt{$X_6$}\ $\land$\ \texttt{$X_7$} \\
$\land$\ \texttt{$X_6$} \texttt{After} \texttt{$X_7$}
} 
& 
\tabincell{l}{
\textcolor{blue}{*} \texttt{$Y$}\ $\gets$\ \texttt{$X_1$}\ $\land$\ \texttt{$X_2$} \\ 
$\land$\ \texttt{$X_3$}  \\
\textcolor{blue}{*} \texttt{$Y$}\ $\gets$\ \texttt{$X_4$}\ $\land$\ \texttt{$X_5$} \\

\textcolor{blue}{*} \texttt{$Y$}\ $\gets$\ \texttt{$X_6$}\ $\land$\ \texttt{$X_8$} \\
\textcolor{blue}{*} \texttt{$Y$}\ $\gets$\ \texttt{$X_7$}\ $\land$\ \texttt{$X_9$} \\ 
$\land$\ \texttt{$X_{10}$} \\
}
\\ \hline

Case-1 & 
\tabincell{l}{
\textcolor{blue}{*} \texttt{$Y$}\ $\gets$\ \texttt{$X_1$}\ $\land$\ \texttt{$X_2$} \\ 
$\land$\ \texttt{$X_3$} \\
$\land$\ \texttt{$X_1$} \texttt{Before} \texttt{$X_2$} 
} & 
\tabincell{l}{
\textcolor{blue}{*} \texttt{$Y$}\ $\gets$\ \texttt{$X_1$}\ $\land$\ \texttt{$X_2$} \\ 
$\land$\ \texttt{$X_3$} \\
$\land$\ \texttt{$X_1$}\ \texttt{Before}\ \texttt{$X_2$} \\
\textcolor{blue}{*} \texttt{$Y$}\ $\gets$\ \texttt{$X_4$}\ $\land$\ \texttt{$X_5$} \\$\land$\ \texttt{$X_4$}\ \texttt{After}\ \texttt{$X_5$} 
} & 
\tabincell{l}{
\textcolor{blue}{*} \texttt{$Y$}\ $\gets$\ \texttt{$X_1$}\ $\land$\ \texttt{$X_2$} \\ 
$\land$\ \texttt{$X_3$}  \\

\textcolor{red}{*} \texttt{$Y$}\ $\gets$\ \texttt{$X_4$}\ $\land$\ \texttt{$X_5$} \\
$\land$\ \texttt{$X_6$} \\
$\land$\ \texttt{$X_4$} \texttt{After} \texttt{$X_6$}\\
$\land$\ \texttt{$X_5$} \texttt{After} \texttt{$X_6$}\\
\textcolor{red}{*} \texttt{$Y$}\ $\gets$\ \texttt{$X_5$}\ $\land$\ \texttt{$X_6$}\\ $\land$\ \texttt{$X_7$} \\
$\land$\ \texttt{$X_5$} \texttt{After} \texttt{$X_6$} \\
$\land$\ \texttt{$X_5$} \texttt{After} \texttt{$X_7$}\\
$\land$\ \texttt{$X_6$} \texttt{After} \texttt{$X_7$}
}

&
\tabincell{l}{
\textcolor{blue}{*} \texttt{$Y$}\ $\gets$\ \texttt{$X_1$}\ $\land$\ \texttt{$X_2$} \\  
$\land$\ \texttt{$X_3$}  \\
\textcolor{red}{*} \texttt{$Y$}\ $\gets$\ \texttt{$X_4$}\ $\land$\ \texttt{$X_5$} \\
$\land$\ \texttt{$X_{10}$} \\

\textcolor{red}{*} \texttt{$Y$}\ $\gets$\ \texttt{$X_6$}\ $\land$\ \texttt{$X_8$} \\
 $\land$\ \texttt{$X_9$}\\
\textcolor{red}{*} \texttt{$Y$}\ $\gets$\ \texttt{$X_8$}\ $\land$\ \texttt{$X_9$} \\ 
$\land$\ \texttt{$X_{10}$} \\
}
\\ \hline

Case-2 & 
\tabincell{l}{
\textcolor{blue}{*} \texttt{$Y$}\ $\gets$\ \texttt{$X_1$}\ $\land$\ \texttt{$X_2$} \\ 
$\land$\ \texttt{$X_3$} \\
$\land$\ \texttt{$X_1$} \texttt{Before} \texttt{$X_2$} 
} & 
\tabincell{l}{
\textcolor{blue}{*} \texttt{$Y$}\ $\gets$\ \texttt{$X_1$}\ $\land$\ \texttt{$X_2$} \\ 
$\land$\ \texttt{$X_3$}  \\

\textcolor{blue}{*} \texttt{$Y$}\ $\gets$\ \texttt{$X_4$}\ $\land$\ \texttt{$X_5$} \\
$\land$\ \texttt{$X_4$} \texttt{Before} \texttt{$X_5$}\\
\textcolor{blue}{*} \texttt{$Y$}\ $\gets$\ \texttt{$X_6$}\ $\land$\ \texttt{$X_7$} \\
$\land$\ \texttt{$X_6$} \texttt{After} \texttt{$X_7$}
} & 
\tabincell{l}{
\textcolor{blue}{*} \texttt{$Y$}\ $\gets$\ \texttt{$X_1$}\ $\land$\ \texttt{$X_2$} \\ 
$\land$\ \texttt{$X_3$}  \\

\textcolor{red}{*} \texttt{$Y$}\ $\gets$\ \texttt{$X_4$}\ $\land$\ \texttt{$X_5$} \\
$\land$\ \texttt{$X_7$}\\
$\land$\ \texttt{$X_4$} \texttt{After} \texttt{$X_7$}\\
$\land$\ \texttt{$X_5$} \texttt{After} \texttt{$X_7$}\\
\textcolor{blue}{*} \texttt{$Y$}\ $\gets$\ \texttt{$X_6$}\ $\land$\ \texttt{$X_7$} \\
$\land$\ \texttt{$X_6$} \texttt{After} \texttt{$X_7$}
}
& 
\tabincell{l}{
\textcolor{blue}{*} \texttt{$Y$}\ $\gets$\ \texttt{$X_1$}\ $\land$\ \texttt{$X_2$} \\ 
$\land$\ \texttt{$X_3$}  \\
\textcolor{red}{*} \texttt{$Y$}\ $\gets$\ \texttt{$X_4$}\ $\land$\ \texttt{$X_5$} \\
$\land$\ \texttt{$X_{8}$} \\

\textcolor{blue}{*} \texttt{$Y$}\ $\gets$\ \texttt{$X_6$}\ $\land$\ \texttt{$X_8$} \\
\textcolor{red}{*} \texttt{$Y$}\ $\gets$\ \texttt{$X_7$}\ $\land$\ \texttt{$X_9$} \\ 
$\land$\ \texttt{$X_{10}$} \\
}
\\ \hline


Case-3 & 
\tabincell{l}{
\textcolor{blue}{*} \texttt{$Y$}\ $\gets$\ \texttt{$X_1$}\ $\land$\ \texttt{$X_2$} \\ 
$\land$\ \texttt{$X_3$} \\
$\land$\ \texttt{$X_1$} \texttt{Before} \texttt{$X_2$} 
} & 
\tabincell{l}{
\textcolor{blue}{*} \texttt{$Y$}\ $\gets$\ \texttt{$X_1$}\ $\land$\ \texttt{$X_2$} \\ 
$\land$\ \texttt{$X_3$} \\
$\land$\ \texttt{$X_1$}\ \texttt{Before}\ \texttt{$X_2$} \\
\textcolor{blue}{*} \texttt{$Y$}\ $\gets$\ \texttt{$X_4$}\ $\land$\ \texttt{$X_5$} \\$\land$\ \texttt{$X_4$}\ \texttt{After}\ \texttt{$X_5$} 
} & 
\tabincell{l}{
\textcolor{blue}{*} \texttt{$Y$}\ $\gets$\ \texttt{$X_1$}\ $\land$\ \texttt{$X_2$} \\ 
$\land$\ \texttt{$X_3$}  \\

\textcolor{blue}{*} \texttt{$Y$}\ $\gets$\ \texttt{$X_4$}\ $\land$\ \texttt{$X_5$} \\
$\land$\ \texttt{$X_4$} \texttt{Before} \texttt{$X_5$}\\
\textcolor{red}{*} \texttt{$Y$}\ $\gets$\ \texttt{$X_9$}\ $\land$\ \texttt{$X_{10}$} \\
$\land$\ \texttt{$X_{11}$} \\
$\land$\ \texttt{$X_9$} \texttt{Equal} \texttt{$X_{11}$} \\
} 
&
\tabincell{l}{
\textcolor{blue}{*} \texttt{$Y$}\ $\gets$\ \texttt{$X_1$}\ $\land$\ \texttt{$X_2$} \\ 
$\land$\ \texttt{$X_3$}  \\
\textcolor{red}{*} \texttt{$Y$}\ $\gets$\ \texttt{$X_4$}\ $\land$\ \texttt{$X_5$} \\
$\land$\ \texttt{$X_{8}$} \\

\textcolor{blue}{*} \texttt{$Y$}\ $\gets$\ \texttt{$X_6$}\ $\land$\ \texttt{$X_8$} \\
\textcolor{red}{*} \texttt{$Y$}\ $\gets$\ \texttt{$X_7$}\ $\land$\ \texttt{$X_9$} \\ 
$\land$\ \texttt{$X_{10}$} \\
$\land$\ \texttt{$X_{7}$} \texttt{After} \texttt{$X_{9}$}
}
\\ \hline

Case-4 & 
\tabincell{l}{
\textcolor{blue}{*} \texttt{$Y$}\ $\gets$\ \texttt{$X_1$}\ $\land$\ \texttt{$X_2$} \\ 
$\land$\ \texttt{$X_3$} \\
$\land$\ \texttt{$X_1$} \texttt{Before} \texttt{$X_2$} 
} & 
\tabincell{l}{
\textcolor{blue}{*} \texttt{$Y$}\ $\gets$\ \texttt{$X_1$}\ $\land$\ \texttt{$X_2$} \\ 
$\land$\ \texttt{$X_3$} \\
$\land$\ \texttt{$X_1$}\ \texttt{Before}\ \texttt{$X_2$} \\
\textcolor{red}{*} \texttt{$Y$}\ $\gets$\ \texttt{$X_2$} $\land$ \texttt{$X_4$}\\
$\land$\ \texttt{$X_5$} \\
$\land$\ \texttt{$X_2$}\ \texttt{Equal}\ \texttt{$X_5$} \\
$\land$\ \texttt{$X_2$}\ \texttt{Equal}\ \texttt{$X_6$} \\
$\land$\ \texttt{$X_5$}\ \texttt{After}\ \texttt{$X_6$} 
} & 
\tabincell{l}{
\textcolor{blue}{*} \texttt{$Y$}\ $\gets$\ \texttt{$X_1$}\ $\land$\ \texttt{$X_2$} \\ 
$\land$\ \texttt{$X_3$}  \\

\textcolor{red}{*} \texttt{$Y$}\ $\gets$\ \texttt{$X_1$}\ $\land$\ \texttt{$X_4$} \\
 $\land$\ \texttt{$X_5$} \\
$\land$\ \texttt{$X_4$} \texttt{Before} \texttt{$X_5$} \\
\textcolor{red}{*} \texttt{$Y$}\ $\gets$\ \texttt{$X_1$}\ $\land$\ \texttt{$X_6$} \\ $\land$\ \texttt{$X_7$} \\
$\land$\ \texttt{$X_6$} \texttt{After} \texttt{$X_7$}
}
&
\tabincell{l}{
\textcolor{blue}{*} \texttt{$Y$}\ $\gets$\ \texttt{$X_1$}\ $\land$\ \texttt{$X_2$} \\ 
$\land$\ \texttt{$X_3$}  \\
\textcolor{blue}{*} \texttt{$Y$}\ $\gets$\ \texttt{$X_4$}\ $\land$\ \texttt{$X_5$} \\

\textcolor{blue}{*} \texttt{$Y$}\ $\gets$\ \texttt{$X_6$}\ $\land$\ \texttt{$X_8$} \\
 $\land$\ \texttt{$X_9$}\\
\textcolor{red}{*} \texttt{$Y$}\ $\gets$\ \texttt{$X_8$}\ $\land$\ \texttt{$X_9$} \\ 
$\land$\ \texttt{$X_{10}$} \\
$\land$\ \texttt{$X_9$} 
\texttt{After} \texttt{$X_{10}$}
\\
}
\\ \hline

Case-5 & 
\tabincell{l}{
\textcolor{blue}{*} \texttt{$Y$}\ $\gets$\ \texttt{$X_1$}\ $\land$\ \texttt{$X_2$} \\ 
$\land$\ \texttt{$X_3$}\\
$\land$\ \texttt{$X_1$} \texttt{Before} \texttt{$X_2$} 
} & 
\tabincell{l}{
\textcolor{blue}{*} \texttt{$Y$}\ $\gets$\ \texttt{$X_1$}\ $\land$\ \texttt{$X_2$} \\ 
$\land$\ \texttt{$X_3$} \\
$\land$\ \texttt{$X_1$}\ \texttt{Before}\ \texttt{$X_2$} \\
\textcolor{red}{*} \texttt{$Y$}\ $\gets$\ \texttt{$X_2$} $\land$ \texttt{$X_4$}\\
$\land$\ \texttt{$X_5$} \\
$\land$\ \texttt{$X_2$}\ \texttt{Equal}\ \texttt{$X_5$} \\
$\land$\ \texttt{$X_2$}\ \texttt{Equal}\ \texttt{$X_6$} \\
$\land$\ \texttt{$X_5$}\ \texttt{After}\ \texttt{$X_6$} 
} &  
\tabincell{l}{
\textcolor{blue}{*} \texttt{$Y$}\ $\gets$\ \texttt{$X_1$}\ $\land$\ \texttt{$X_2$} \\ 
$\land$\ \texttt{$X_3$}  \\

\textcolor{red}{*} \texttt{$Y$}\ $\gets$\ \texttt{$X_1$}\ $\land$\ \texttt{$X_4$} \\
 $\land$\ \texttt{$X_5$} \\
$\land$\ \texttt{$X_4$} \texttt{Before} \texttt{$X_5$} \\
\textcolor{red}{*} \texttt{$Y$}\ $\gets$\ \texttt{$X_1$}\ $\land$\ \texttt{$X_6$} \\ $\land$\ \texttt{$X_7$} \\
$\land$\ \texttt{$X_6$} \texttt{After} \texttt{$X_7$}
}
& 
\tabincell{l}{
\textcolor{blue}{*} \texttt{$Y$}\ $\gets$\ \texttt{$X_1$}\ $\land$\ \texttt{$X_2$} \\ 
$\land$\ \texttt{$X_3$}  \\
\textcolor{red}{*} \texttt{$Y$}\ $\gets$\ \texttt{$X_4$}\ $\land$\ \texttt{$X_5$} \\
$\land$\ \texttt{$X_{10}$} \\

\textcolor{blue}{*} \texttt{$Y$}\ $\gets$\ \texttt{$X_6$}\ $\land$\ \texttt{$X_8$} \\
 $\land$\ \texttt{$X_9$}\\
\textcolor{red}{*} \texttt{$Y$}\ $\gets$\ \texttt{$X_8$}\ $\land$\ \texttt{$X_9$} \\ 
$\land$\ \texttt{$X_{10}$} \\
}
\\ \hline

\end{tabular}
\caption{Results of Group-1, Group-2, Group-3 and Group-4. \textcolor{blue}{*}Ground truth rules / learned grounded rules. \textcolor{red}{*} Rules that are wrongly learned. \textcolor{green}{*} Rules that are not learned. }  
\label{appendix_tab:all_groups}
\end{table}

\textbf{Scalability of Our Method}  The scalability results are illustrated in Fig.\ref{accuracy_and_time}, where we report how the rule learning accuracy and the computation time change over different predicate sets and sample size. As the sample sizes increase, the accuracy of learned temporal logic rules improves, but at the same time, the required training time also increases. When the number of ground truth temporal logic rules increases, there is a slight decrease in learning accuracy, but it remains within an acceptable range. From this result, we conclude that our method scales fairly well with the growing number of predicates and sample sizes.

\clearpage

\clearpage

\section{Computing Infrastructure} 
\label{sec:computing_infrastructure}
All synthetic data experiments, as well as the real-world data experiments, including the comparison experiments with baselines,  are performed on Ubuntu 20.04.3 LTS system with Intel(R) Xeon(R) Gold 6248R CPU @ 3.00GHz, 227 Gigabyte memory.

\section{About Baselines}
\label{sec:baselines}
We consider the following baselines through synthetic data experiments and healthcare data experiments to compare the rule learning ability and event prediction with our proposed model:

\textbf{Neural-based (black-box) models for irregular event data}

\begin{itemize}
\item Transformer Hawkes Process (THP)~\citep{zuo2020transformer}: It is a concurrent self-attention based point process model with additional structural knowledge.
\item Recurrent Marked Temporal Point Processes (RMTPP) ~\citep{du2016recurrent}: It uses RNN to learn a representation of the past events and time intervals.
\item ERPP~\citep{xiao2017modeling}: It uses two RNN models for the background of intensity function and the history-dependent intensity part. 
\item LG-NPP algorithm ~\citep{zhang2021learning}: It applies self-attention to embed the event sequence data. It uses a latent random graph to model the relationship between different event sequences and proposes a bilevel programming algorithm to uncover the latent graph and embedding parameters.
\end{itemize}

\textbf{Simple parametric/nonparametric models for irregular event data}
\begin{itemize}
\item Granger Causal Hawkes (GCH) ~\citep{xu2016learning}: A multivariate Hawkes model with sparse-group-lasso and pairwise similarity constriants. It uses an EM learning algorithm with the basis function of Hawkes process selected adaptively. 
\item GM-NLF algorithm ~\citep{eichler2017graphical}: It is a  multivariate Hawkes Processes with nonparameteric intensity function.
\end{itemize}

\textbf{Logical models for irregular event data}
\begin{itemize}
\item Clock Logic Neural Networks (CLNN) ~\citep{yan2023weighted}: It learns weighted clock logic (wCL) formulas as interpretable temporal rules by which some events promote or inhibit other events. Moreover, CLNN models temporal relations between events using conditional intensity rates informed by a set of wCL formulas.
\item TELLER ~\citep{li2021explaining}: A non-differentiable algorithm for uncovering temporal logic rules.
\end{itemize}

We compared our model with these baselines in event prediction accuracy. Our model achieves promising results and meanwhile is interpretable as CLNN.

\textbf{Itemset mining methods}
\begin{itemize}
\item Apriori ~\citep{agrawal1994fast}: An itemset mining method, which identifies frequent individual items and iteratively extends them to larger itemsets.
\item NEclatcloesed ~\citep{aryabarzan2021neclatclosed}: An itemset mining method, which enhances Apriori's speed and memory efficiency, by employing a depth-first traversal of an itemset search tree, constructing partial closures, and using a hashmap for subsumption checking.
\end{itemize}

\textbf{Sequential pattern mining}
\begin{itemize}
\item CM-spade ~\citep{fournier2014fast}: A sequential pattern mining method, which employs co-occurrence data via a compact Co-occurrence Map (CMAP) for pruning infrequent candidates. CM-SPADE integrates CMAP into the SPADE algorithm, enhancing performance by eliminating costly database scans.
\item VGEN ~\citep{fournier2014vgen}: A sequential pattern mining method, extracts fewer yet more representative sequential patterns using a depth-first search approach and leverages co-occurrence data for candidate pruning.
\end{itemize}

\textbf{Sequential rule mining}
\begin{itemize}
\item ERMiner ~\citep{fournier2014erminer}: A sequential rule mining method, which extracts sequential rules from sequences, utilizing equivalence classes, merging strategies, and a Sparse Count Matrix for efficient search and candidate pruning. 
\end{itemize}

\textbf{Inductive logic programming (ILP)}
\begin{itemize}
\item Aleph ~\citep{srinivasan2001aleph}: Utilizes a general-to-specific hill-climbing strategy to derive logical theories from positive and negative examples. It starts with a broad clause encompassing all positives, iteratively refining it by adding literals to maintain positive coverage and exclude negatives. Note that ILR is a supervised rule learning method. To implement ILR, we need to release the event label information to the algorithm, which poses an unfair comparision with our method. 
\end{itemize}

We compared our algorithm with these baselines in (temporal) rule mining accuracy with the results summarized and compared in Fig.\ref{other_rule_mining_method}. We also explained the performance metric that can be used to evaluate the rule learning accuracy.

\textbf{Post-hoc method}
\begin{itemize}
\item Dynamask ~\citep{crabbe2021explaining}: A SOTA post-hoc method "Dynamask" that aims to explain the black-box time series models by learning dynamic masks.
\end{itemize}

We compared the interpretability performance of this post-hoc method with ours on the MIMIC dataset and found medical references to justify the results. Details can be found in Sec.~\ref{sec:post-hoc}.

\section{Comparison with post-hoc methods}
\label{sec:post-hoc}
Post-hoc methods are techniques that try to explain the predictions of a deep learning model after the model is well  trained. It is easy to implement. However, the reliability and applicability of post-hoc method have always been questioned~\citep{rudin2019stop}. One key concern is that the post-hoc methods explain the models not the world. 

In contrast, logic-based models possess inherent interpretability and function as transparent white-box solutions, although the computation is more challenging. 

To further support our arguments, we adopted a SOTA post-hoc method Dynamask ~\citep{crabbe2021explaining} that aims to explain the black-box time series models by learning dynamic masks. We compare this post-hoc method with our method in terms of knowledge discovery. 

We apply Dynamask to our MIMIC-IV datasets. The method assumption is that we have trained a flexible black-box model (such as LSTM) that can accurately capture the complex dynamics of multivariate time series. To explain the black-box model in a post-hoc way, we will construct dynamic perturbation operators, with the mask coefficients indicating the saliency of each feature at each time step.

We considered a single layer LSTM with 128 hidden states as our black-box model and fed sufficient time series sequences for training. These sequences included lab values in numerical form, queried from the MIMIC-IV dataset. In the subsequent post-hoc explanation phase, we employed the Dynamask approach to acquire interpretable sparse binary dynamic masks. These masks were trained to provide explanations, focusing on segments of time series spanning a 48-hour time window containing instances of abnormal low urine output.

The goal here is to provide counterfactual explanations, such as determining how historical attributes could have been altered to avert occurrences of low urine output events.

In the below table, we list several representative discovered patterns in the learned masks. For each pattern, we report some important ($\text{event}_i$, $\text{time}_j$), observational status, and corresponding counterfactual status, that is, when the observation of event $i$ at time $j$ is perturbed by using the observation of this event at adjacent times, it will cause the original abnormal "Low Urine Output" in the 48th hour to change to the "Normal Higher Value".

\begin{table}[ht]
\label{appendix_tab:dynamask}
\begin{tabular}{lllll}
\hline
\textbf{Patient-ID} & \textbf{Time} & \textbf{Event} & \textbf{Observational Status} & \textbf{Counterfactual Status} \\ \hline
\multirow{6}{*}{10606611} & 17 & Arterial Blood Pressure systolic & abnormal & normal \\ \cline{2-5} 
 & 17 & Arterial Blood Pressure mean & abnormal & normal \\ \cline{2-5} 
 & 17 & Respiratory Rate & abnormal & normal \\ \cline{2-5} 
 & 11 & Venous O2 Pressure & abnormal & normal \\ \cline{2-5} 
 & 22 & Venous O2 Pressure & abnormal & normal \\ \cline{2-5} 
 & 28 & Venous O2 Pressure & abnormal & normal \\ \hline
\multirow{6}{*}{10623984} & 42 & Sodium & normal & abnormal \\ \cline{2-5} 
 & 42 & Chloride & normal & abnormal \\ \cline{2-5} 
 & 42 & Hemoglobin & abnormal & normal \\ \cline{2-5} 
 & 42 & Platelet Count & abnormal & normal \\ \cline{2-5} 
 & 39 & Lactic Acid & abnormal & normal \\ \cline{2-5} 
 & 46 & Lactic Acid & abnormal & normal \\ \hline
\multirow{6}{*}{11281568} & 47 & SpO2 & abnormal & normal \\ \cline{2-5} 
 & 3 & Sodium & normal & abnormal \\ \cline{2-5} 
 & 30 & Sodium & normal & abnormal \\ \cline{2-5} 
 & 34 & Sodium & normal & abnormal \\ \cline{2-5} 
 & 32 & WBC & abnormal & normal \\ \cline{2-5} 
 & 40 & WBC & abnormal & normal \\ \hline
\multirow{7}{*}{11395953} & 37 & Sodium & normal & abnormal \\ \cline{2-5} 
 & 37 & BUN & abnormal & normal \\ \cline{2-5} 
 & 37 & IRN & normal & abnormal \\ \cline{2-5} 
 & 37 & Venous O2 Pressure & abnormal & normal \\ \cline{2-5} 
 & 37 & Venous O2 Pressure & abnormal & normal \\ \cline{2-5} 
 & 45 & Lactic Acid & abnormal & normal \\ \cline{2-5} 
 & 47 & Lactic Acid & abnormal & normal \\ \hline
\end{tabular}

\caption{Representative Patterns derived using Dynamask.}
\end{table}

The results may imply that abnormal status of variables such as Arterial Blood Pressure, Arterial CO2 Pressure, Venous O2 Pressure, Respiratory Rate, and others play a crucial role in influencing changes in patients' urine output. These findings align with the discoveries outlined in our paper. Additionally, sudden changes in Lactic Acid and Platelet Count should also alert patients.

Some of Dynamask's conclusions are consistent with our findings. However, unlike our method, which has been strictly reviewed and affirmed by medical experts, Dynamask's findings have not been confirmed by experts. Furthermore, we verify and compare some inaccuracies in counterfactual interpretation with reference to the medical literature. Experiment results have indicated that Dynamask might overlook critical clinical features.

For instance, it omitted tachycardia (abnormal heart rate), a significant omission. Tachycardia is a well-established early symptom of sepsis, observed in over 80\% of patients~\citep{komorowski2018artificial}. Tachycardia reflects compensatory mechanisms responding to infection-induced hypovolemia and myocardial depression, both influencing urine output~\citep{singer2016third}. In contrast, our method effectively incorporates abnormalities in heart rate within learned rules, such as

{\bf Rule 6: LowUrine $\leftarrow$ abnormal Heart Rate $\wedge$ abnormal BUN $\wedge$ abnormal WBC, abnormal Heart Rate Before abnormal BUN, abnormal BUN Before abnormal WBC}

Experiment results have also shown that Dynamask may discover patterns that contradict clinical knowledge. For instance, clinical understanding suggests that some sepsis patients may experience extracellular fluid volume depletion, leading to hyponatremia~\citep{bagshaw2009disorders}. Sepsis-induced inflammatory response and fluid shifts can decrease blood sodium concentration, potentially causing concentrated urine and reduced urine output. Only when blood sodium concentration returns to normal from an abnormal state might urine output normalize. However, Dynamask's patterns for specific patients exhibit perplexing opposite scenarios, diminishing the algorithm's reliability.

The above examples reflect that our method better aligns with established clinical understanding of sepsis, whereas Dynamask occasionally fails to capture the intricate physiological interplay between sepsis, fluid/electrolyte status, and urine output.

In summary, our logic-based has the following advantages over Dynamask (and maybe other post-hoc methods):

(1) Post-hoc methods rely on accurate black-box models, which are hard to obtain when the sample size is small.

(2) The instance-wise explanations generated by Dynamask are not enough to provide a global understanding of the learned model. Our method can easily generalize through learning logical rules and be used in inductive settings.

(3) Dynamask's explanations can only reflect the importance of each feature independently and cannot express the relationships between features, requiring further human interpretation. Our method can directly learn interpretable rules and their corresponding weights.

\section{Compare with other rule mining methods}
\label{sec:compare_with_other_baselines}
\begin{figure}[ht]
\centering
\includegraphics[width=1\textwidth]{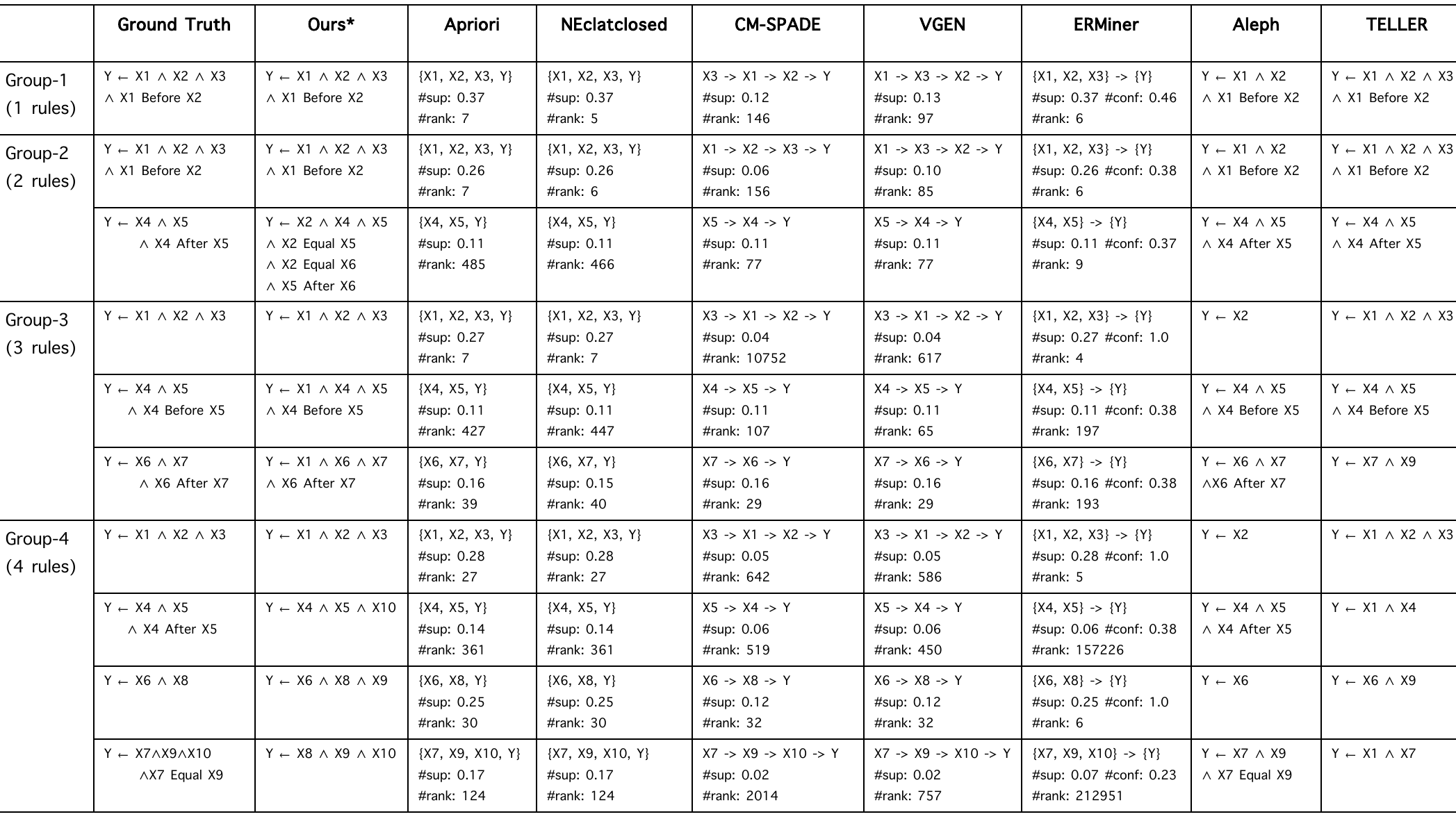}
\caption{Learned rules using newly added baselines. $\# sup$ is support, the number of samples that contain corresponding pattern divided by the number of all samples. $\# conf$ is confidence, the number of samples that contain all items of $X$ before all items of $Y$, divided by the number of samples that contains items in $X$. $\# rank$ is the descending order of $\# conf$ for a pattern among all $Y$-related patterns. }
\label{other_rule_mining_method}
\end{figure}

Association Rule Mining (ARM) is an unsupervised learning algorithm aimed at identifying frequent itemsets using metrics like support and confidence that are then utilized to deduce association rules. 

As shown in Fig.\ref{other_rule_mining_method}, we compare three types of ARM: \textbf{itemset mining methods}, which ignore event ordering information, including Apriori~\citep{agrawal1994fast} and NEclatcloesed~\citep{aryabarzan2021neclatclosed}; \textbf{sequential pattern mining}, which cannot handle precise timestamps, including CM-spade~\citep{fournier2014fast} and VGEN~\citep{fournier2014vgen}; and \textbf{sequential rule mining}, which ignore internal temporal relations, including ERMiner~\citep{fournier2014erminer}. The learned rules by itemset mining methods like $\left \{ X,Y \right \}$ only capture event co-occurrence, neglecting temporal relationships. The acquired rules by sequential pattern mining method take the form of $X_i\rightarrow...\rightarrow X_j\rightarrow Y$, which represents sequences where events strictly adhere to a defined temporal relationship. Learned rules by sequential rule mining method are like $\left \{ X \right \} \rightarrow \left \{ Y \right \}$, where events in the rule body precede those in the rule head, but internal temporal relations within the body and head events are disregarded. Unlike ARM, our method considers a range of temporal relations (Before, After, Equal, None) between pairwise body predicates. Our rule forms encompass ARM, enabling us to manage more intricate temporal relationships.

Further, we consider \textbf{inductive logic programming (ILP) method}, which is a supervised technique that learns logical rules from factual data, positive and negative examples. It prioritizes positive coverage while minimizing negative coverage and can incorporate background knowledge through extended logic programs. ILP requires labeled positive and negative samples to learn logic rules, which can be expensive or hard to acquire in practice. To implement ILR in our setting, we release the true event label information to ILR, which poses an unfair comparison to our method. In contrast, our algorithm only utilizes the event items' timestamps or ordering information for rule learning, treating label information as a latent variable inferred in the E-step and thereby does not require prior knowledge of the sample labels.

A rigorous comparison between these baselines and our method is shown in Tab.\ref{tab:rigorous_comparison_with_baselines}, where we selected three representative ARM methods to summarize the results for Group-1 to Group-4 datasets below for a clear comparison. The metrics we used are as follows:

\begin{itemize}
    \item \textbf{$\#$Support}: is the threshold hyperparameter for baseline methods, determining the rule appearance frequency threshold of the discovered rules in the final set.
    \item \textbf{$\#$Rank}: represents the occurrence frequency ranking of true rules in the discovered rule set. A high $\#$Rank indicates practical unfeasibility
    \item \textbf{Jaccard index}: (between 0 and 1) measures overlap between ground truth rule set and the discovered rule set, counting shared and distinct members.
\end{itemize}

\begin{table}[ht]
\centering
\begin{tabular}{c|c|c|c|c} 
\toprule
Group & Method & $\#$support & $\#$rank & Jaccard Index \\ 
\hline
Group-1 & Apriori & 0.37 & 7 & 0\\
        & CM-SPADE & 0.12 & 146 & 0\\
        & ERMiner & 0.37 & 6 & 0\\
        & \textbf{Ours*} & -- & -- & 1 \\
\hline
Group-2 & Apriori & 0.26, 0.11  & 7, 485 & 0 \\
        & CM-SPADE & 0.11, 0.06 & 77, 156 & 0 \\
        & ERMiner & 0.26, 0.11 & 6, 9 & 0 \\
        & \textbf{Ours*} & -- & -- & 0.5 \\
\hline
Group-3 & Apriori & 0.27, 0.16, 0.11 & 7, 39, 427 & 0 \\
        & CM-SPADE & 0.16, 0.11, 0.04 & 29, 107, 10752 & 0 \\
        & ERMiner & 0.27, 0.16, 0.11 & 4, 193, 197 & 0 \\
        & \textbf{Ours*} & -- & -- & 0.33 \\
\hline
Group-4 & Apriori & 0.28, 0.25, 0.17, 0.14 & 27, 30, 124, 361 & 0 \\
        & CM-SPADE & 0.12, 0.06, 0.05, 0.02 & 32, 519, 642, 2014 & 0 \\
        & ERMiner & 0.28, 0.25, 0.07, 0.06 & 5, 6, 157226, 212951 & 0 \\
        & \textbf{Ours*} & -- & -- & 0.25 \\
\bottomrule
\end{tabular}

\caption{Compare our method with other rule mining baselines using $\#$support, $\#$rank, and jaccard similarity metircs.}
\label{tab:rigorous_comparison_with_baselines}
\end{table}

Tab.\ref{tab:rigorous_comparison_with_baselines} confirms our method's better rule discovery accuracy compared to these rule mining baselines. Significantly, despite the increased problem difficulty from Group-1 to Group-4, our method's Jaccard index drops, our discovered rules still consistently capture the majority or partial rule content. While baseline methods may include true rules in their discovered sets, the high $\#$Rank suggests practical challenges in extracting them from such extensive rule sets.

\section{MIMIC-IV dataset preprocessing and risk factors extracting}
\label{sec:variables_describe}

MIMIC-IV\footnote{\url{https://mimic.mit.edu/}} is a publicly available database sourced from the electronic health record of the Beth Israel Deaconess Medical Center~\citep{johnson2023mimic}. Information available includes patient measurements, orders, diagnoses, procedures, treatments, and deidentified free-text clinical notes. Sepsis is a leading cause of mortality in the ICU, particularly when it progresses to septic shock. Septic shocks are critical medical emergencies, and timely recognition and treatment are crucial for improving survival rates. In the real-world experiments on MIMIC-IV dataset, we aim to find logic rules related to septic shocks for the whole patient samples and infer the most likely rule reasons for specific patients, which would be potential early alarm when some abnormal indicators occur.

\textbf{Patients} We select 4074 patients that satisfied the following criteria from the dataset: (1) The patients are diagnosed with sepsis~\citep{saria2018individualized}. (2) Patients, if diagonized with sepsis, the timestamps of any clinical testing, specific lab values, timestamps of medication administration and corresponding dosage were not missing.

\textbf{Outcome} Real time urine output was treated as the outcome indicator since low urine output signals directly indicate a poor circulatory system and is a warning sign of septic shock. 

\textbf{Risk Factors} Suggested by \citep{komorowski2018artificial}, we extract 28 risk factors associated with sepsis which are consistent with expert consensus. Based on the distinct clinical characteristics of these risk factors, they can be categorized into the following five groups:

\begin{itemize}
    \item Vital Signs:
    \begin{itemize}
        \item Heart Rate: The number of times the heart beats per minute. An elevated or abnormal heart rate may indicate physiological stress or an underlying condition.
        \item Arterial Blood Pressure (systolic, mean, diastolic): Measures the force exerted by the blood against the arterial walls during different phases of the cardiac cycle. Abnormal blood pressure values may indicate cardiovascular dysfunction or organ perfusion issues.
        \item Temperature (Celsius): Body temperature is a measure of the body's internal heat. Abnormal temperatures may indicate infection, inflammation, or other systemic disorders.
        \item Respiratory Rate: The number of breaths taken per minute. Abnormal respiratory rates may suggest respiratory distress or dysfunction.
        \item $SpO_2$: Oxygen saturation level in the blood. Decreased $SpO_2$ levels may indicate inadequate oxygenation.
    \end{itemize}
    
    \item Biochemical Parameters:
    \begin{itemize}
        \item Potassium, Sodium, Chloride, Glucose: Electrolytes and blood sugar levels that help maintain essential bodily functions. Abnormal levels may indicate electrolyte imbalances, metabolic disorders, or organ dysfunction.
        \item Blood Urea Nitrogen (BUN), Creatinine: Indicators of renal function. Elevated levels may suggest impaired kidney function.
        \item Magnesium, Ionized Calcium: Important minerals involved in various physiological processes. Abnormal levels may indicate electrolyte imbalances or organ dysfunction.
        \item Total Bilirubin: A byproduct of red blood cell breakdown. Elevated levels may indicate liver dysfunction.
        \item Albumin: A protein produced by the liver. Abnormal levels may indicate malnutrition, liver disease, or kidney dysfunction.
    \end{itemize}

    \item Hematological Parameters
    \begin{itemize}
        \item Hemoglobin: A protein in red blood cells that carries oxygen. Abnormal levels may indicate anemia or oxygen-carrying capacity issues.
        \item White Blood Cell (WBC): Cells of the immune system involved in fighting infections. Abnormal levels may indicate infection or inflammation.
        \item Platelet Count: Blood cells responsible for clotting. Abnormal levels may suggest bleeding disorders or impaired clotting ability.
        \item Partial Thromboplastin Time (PTT), Prothrombin time (PT), INR: Tests that assess blood clotting function. Abnormal results may indicate bleeding disorders or coagulation abnormalities.
    \end{itemize}

    \item Blood Gas Analysis
    \begin{itemize}
        \item pH (Arterial): A measure of blood acidity or alkalinity. Abnormal pH values may indicate acid-base imbalances or respiratory/metabolic disorders.
        \item Arterial Base Excess: Measures the amount of excess or deficit of base in arterial blood. Abnormal levels may indicate acid-base imbalances or metabolic disturbances.
        \item Arterial CO2 Pressure, Venous O2 Pressure: Parameters that assess respiratory and metabolic function. Abnormal values may indicate respiratory failure or metabolic disturbances.
    \end{itemize}

    \item Metabolic Parameter
    \begin{itemize}
        \item Lactic Acid: An indicator of tissue perfusion and oxygenation. Elevated levels may suggest tissue hypoxia or impaired cellular metabolism.
    \end{itemize}

\end{itemize}

These risk factors are commonly assessed in sepsis patients to monitor their clinical status and guide appropriate interventions. The interpretation of these factors requires clinical judgment and consideration of the patient's overall condition. Tab.\ref{Variables description} shows the categories of the 28 variables extracted from MIMIC-IV dataset, and their reference range.

\textbf{Data Preprocessing} Due to the frequent fluctuations in urine output within the ICU setting, we considered only those instances in which urine output became abnormal after maintaining a normal level for at least 48 hours. These instances were regarded as valid target events that hold significance for prediction and explanation. Additionally, for each patient, we selectively extracted the initial time period that met the criteria. Regarding the risk factors, we documented the time points at which these variables first became abnormal from normal within the 48-hour period preceding the transition of urine output from normal to abnormal.

\begin{table*}[t]
\centering
\begin{tabular}{p{4.5cm}|p{4cm}|p{2.2cm}|p{2cm}}
\hline
\multicolumn{1}{c|}{\textbf{Category}} & \multicolumn{1}{c|}{\textbf{Risk Factor}} & \multicolumn{1}{c|}{\textbf{Ref Range Lower}} & \multicolumn{1}{c}{\textbf{Ref Range Higher}} \\
\hline
\multirow{7}{*}{\textbf{Vital Signs}} & Heart Rate & 60/min & 100/min \\
\cline{2-4}
  & Arterial Blood Pressure systolic & 90 mmHg & 120 mmHg \\
\cline{2-4}
  & Arterial Blood Pressure mean & 70 mmHg & 100 mmHg \\
\cline{2-4}
  & Arterial Blood Pressure diastolic & 60 mmHg & 80 mmHg \\
\cline{2-4}
 & Temperature Celsius & 36.5 & 37.5 \\
\cline{2-4}
  & Respiratory Rate & 12/min & 20/min \\
\cline{2-4}
  & $SpO_2$ & 95$\%$ & - \\

\hline
\multirow{10}{*}{\textbf{Biochemical Parameters}} & Potassium & 3.5 mmol/L & 5.0 mmol/L \\
\cline{2-4}
 & Sodium & 135 mmol/L & 145 mmol/L \\
\cline{2-4}
& Chloride & 98 mmol/L & 106 mmol/L \\
\cline{2-4}
& Glucose & 3.9 mmol/L & 5.6 mmol/L \\
\cline{2-4}
& BUN & 7 mg/dL & 20 mg/dL \\
\cline{2-4}
& Creatinine & 0.5 mg/dL & 1.2 mg/dL \\
\cline{2-4}
& Magnesium & 1.7 mg/dL & 2.3 mg/dL \\
\cline{2-4}
& Ionized Calcium & 4.5 mg/dL & 5.5 mg/dL \\
\cline{2-4}
& Total Bilirubin & - & 1.2 mg/dL \\
\cline{2-4}
& Albumin & 3.4 g/dL & 5.4 g/dL \\

\hline
\multirow{4}{*}{\textbf{Hematological Parameters}} & Hemoglobin & 12.0 g/dL & 17.5 g/dL \\
\cline{2-4}
& WBC & 4000 cells/$\text{mm}^3$ & 11000 cells/$\text{mm}^3$ \\
\cline{2-4}
& Platelet Count & 150000 platelets/$\text{mm}^3$ & 450000 platelets/$\text{mm}^3$ \\
\cline{2-4}
& PTT & 25s & 35s \\
\cline{2-4}
& PT & 11s & 13s \\
\cline{2-4}
& INR & 0.9 & 1.1 \\

\hline
\multirow{4}{*}{\textbf{Blood Gas Analysis}} & pH (Arterial) & 7.35 &  7.45 \\
\cline{2-4}
& Arterial $CO_2$ Pressure & 35 mmHg & 45 mmHg \\
\cline{2-4}
& Venous $O_2$ Pressure & 40 mmHg & - \\
\cline{2-4}
& Arterial Base Excess & -2 mmol/L & 2 mmol/L \\

\hline
\multirow{1}{*}{\textbf{Metabolic Parameter}} & Lactic Acid & - & 2 mmol/L \\

\hline
\end{tabular}
\caption{Description of the risk factors extracted from MIMIC-IV dataset and their corresponding reference range. Variables like $SpO_2$, Total Bilirubin, Venous $O_2$ Pressure, and Lactic Acid only have an upper or lower limit for the reference value.}
\label{Variables description}
\end{table*}

\clearpage

\section{Doctor Verification and Medical References} 
\label{sec:doctor_verification}
Evaluating interpretability of learned temporal logic rules in real-world problems require human expert inspection. In our clinical experiments using the MIMIC-IV dataset, our methodology included the invaluable input of medical experts who reviewed our discovered rule results. Their feedback consistently affirmed the sensibility of our findings. Furthermore, we reinforced our discovered rules with evidence from medical references, as exemplified below:

\begin{itemize}
    \item \textbf{Rule 1: LowUrine $\leftarrow$ Arterial Blood Pressure Diastolic}: Arterial diastolic blood pressure serves as an indicator of renal perfusion pressure. A decrease in arterial diastolic blood pressure can diminish vascular perfusion to the kidneys, impacting the glomerular filtration rate and urine output for sepsis patients ~\citep{gernardin1996blood, benchekroune2008diastolic}. 
    \item \textbf{Rule 2: LowUrine $\leftarrow$ Ionized Calcium}: Decreased ionized calcium levels are common in critically ill sepsis patients due to an inflammatory response that can inhibit calcium absorption ~\citep{mizock1995alterations}. Sepsis-related damage to the gastrointestinal mucosa further reduces dietary calcium absorption ~\citep{zivin2001hypocalcemia}.
    \item \textbf{Rules 3: LowUrine $\leftarrow$ Arterial CO2 Pressure, Rule 4: LowUrine $\leftarrow$ Venous O2 Pressure}: These rules involve arterial CO2 pressure and venous O2 pressure, both of which are linked to cardiac output and tissue hypoxia in septic shock ~\citep{mallat2014central,mecher1990venous,durkin1993relationship,mecher1990venous}.
    \item \textbf{Rule 5: LowUrine $\leftarrow$ Respiratory Rate $\wedge$ Hemoglobin, Respiratory Rate None Hemoglobin}: In early sepsis stages, peripheral vasodilation can lead to hypovolemic shock, decreasing renal perfusion pressure and renal function \citep{levy2001international,langenberg2006renal}. Sepsis-induced anemia exacerbates tissue hypoxia ~\citep{gomez2014unified}.
    \item \textbf{Rule 6: LowUrine $\leftarrow$ Heart Rate $\wedge$ BUN $\wedge$ WBC, Heart Rate Before BUN, BUN Before WBC}: Sequential abnormalities in heart rate, blood urea nitrogen (BUN), and white blood cell (WBC) levels suggest developing renal dysfunction and reduced urine output in sepsis ~\citep{annane2005septic}. These markers indicate sepsis-induced renal hypoperfusion and glomerular filtration decline.
\end{itemize}

*Note that the predicates in the body of the above-mentioned learned rules are all in an abnormal state.

\clearpage
\bibliographystyle{plainnat}

\end{document}